%% file: main.tex
\pdfoutput=1
\documentclass{article}

\PassOptionsToPackage{numbers, compress}{natbib}

\usepackage[main, final]{neurips_2026}

\usepackage[utf8]{inputenc}
\usepackage[T1]{fontenc}
\usepackage{hyperref}
\usepackage{url}
\usepackage{booktabs}
\usepackage{amsfonts}
\usepackage{amsmath}
\usepackage{amssymb}
\usepackage{amsthm}
\usepackage{nicefrac}
\usepackage{microtype}
\usepackage{xcolor}
\usepackage{graphicx}
\graphicspath{{figures/}}
\usepackage{algorithm}
\usepackage{algorithmic}
\usepackage{subcaption}
\usepackage{enumitem}
\usepackage{multirow}
\usepackage{tablefootnote}
\usepackage{tikz}
\usetikzlibrary{arrows.meta, positioning, calc}

\input{common_macros}

\title{Coverage-Verified Sparse Attention: Closed-Loop \\Quality Control for Long-Context LLM Decoding}

\author{%
  Yan-Cheng Li \\
  National Taiwan University \\
  \texttt{r15943142@ntu.edu.tw} \\
  \And
  Chun-Yi Lee \\
  National Taiwan University \\
  \texttt{cylee@csie.ntu.edu.tw} \\
}
\begin{document}

\maketitle

\begin{abstract}
\cyl{%
Sparse attention accelerates long-context decoding by reading a subset
of the KV cache, but existing methods are open-loop: the kernel
commits to its selection with no signal indicating whether enough
attention mass was preserved and no recovery path when it was not.
The needed signal already exists inside the kernel.
The fraction of softmax mass on the selected tokens, which we call
\emph{coverage}, satisfies an exact algebraic equality with per-step
output error and serves a dual role as the acceptance criterion and
the interpolation weight for recovery.
\emph{Coverage-Verified Sparse Attention} (\ssa{}) turns coverage
into a training-free, closed-loop verify-then-correct decode loop that
accepts high-coverage drafts and recovers only the heads that fall
short, with no sampling, no draft model, and no speculative-decoding
infrastructure.
Quality varies by only $0.14$ points across the threshold sweep;
budget and scheduling ablations characterize the deployed operating point.
On LongBench~\citep{bai2024longbench} at 7B and 70B, \ssa{} comes within $1.13$ points of dense attention,
statistically matching it at 70B, and it loses $0.8$ points on RULER;
measured median decode-step speedup reaches $6.21\times$ at 128K context.%
}
\end{abstract}

\input{sections/introduction}

\input{sections/preliminary}

\input{sections/method}

\input{sections/experiments}

\input{sections/related_work}

\input{sections/conclusion}

\begin{ack}
The authors gratefully acknowledge the support from the National Science and Technology Council (NSTC) in Taiwan under grant numbers \mbox{NSTC 114-2221-E-002-069-MY3}, \mbox{NSTC 113-2221-E-002-212-MY3}, \mbox{NSTC 115-2218-E-A49-010}, and \mbox{NSTC 115-2640-E-002-014}, as well as the support from the Academia Sinica Scholar Award (ASSA) under grant number \mbox{AS-ASSA-115-02}, NTU Artificial Intelligence Center of Research Excellence, and Taiwan Centers of Excellence in Artificial Intelligence. This research was also supported by the NVIDIA Academic Grant Program. The authors would also like to express their appreciation for the hardware grant donation of the GPUs from NVIDIA Corporation and NVIDIA AI Technology Center (NVAITC) used in this work. Furthermore, the authors extend their gratitude to the National Center for High-Performance Computing (NCHC) for providing computational and storage resources. The authors also thank the NVIDIA Taipei-1 supercomputer for providing essential computing resources.
\end{ack}

\clearpage
\bibliographystyle{unsrtnat}
\bibliography{references}

\clearpage
\appendix

\begin{center}
  {\Large\bf Appendix}
\end{center}

\input{appendix/algorithm}

\input{appendix/proofs}

\input{appendix/bootstrap}

\input{appendix/longbench}

\input{appendix/m_sweep}

\input{appendix/dense_recovery_ablation}

\input{appendix/budget_fairness}

\input{appendix/coverage_landscape}

\input{appendix/recovery_heatmap}

\input{appendix/ruler}

\input{appendix/llama70b}

\input{appendix/roofline}

\input{appendix/throughput}

\input{appendix/iso_bandwidth}

\input{appendix/selector_traffic}

\input{appendix/limitations}

\input{appendix/reproducibility}

\input{appendix/ruler_tables}

\clearpage

\end{document}

%% file: common_macros.tex
\newtheorem{proposition}{Proposition}

\newcommand{\ssa}{\textsc{CVSA}}
\newcommand{\odense}{O_{\text{dense}}}
\newcommand{\odraft}{O_{\text{draft}}}
\newcommand{\ocomp}{O_{\text{comp}}}
\newcommand{\Sel}{{S_{\text{sel}}}}
\newcommand{\Comp}{{S_{\text{comp}}}}
\newcommand{\todo}[1]{\textcolor{red}{[TODO: #1]}}
\definecolor{CVSAAdd}{RGB}{0,0,0}
\definecolor{CVSATableBest}{RGB}{255,241,181}
\definecolor{CVSATableGood}{RGB}{221,244,241}
\newcommand{\cvsaadd}[1]{#1}
\newcommand{\cvstabbest}[1]{\begingroup\setlength{\fboxsep}{1.1pt}\colorbox{CVSATableBest}{\textbf{#1}}\endgroup}
\newcommand{\cvstabgood}[1]{\begingroup\setlength{\fboxsep}{1.1pt}\colorbox{CVSATableGood}{#1}\endgroup}
\newcommand{\cvsadel}[1]{}

\usepackage{xspace}

\newcommand{\cyl}[1]{#1}

\usepackage[capitalize,noabbrev]{cleveref}
\AtBeginDocument{%
  \crefname{figure}{Fig.}{Figs.}%
  \Crefname{figure}{Fig.}{Figs.}%
  \crefname{table}{Table}{Tables}%
  \Crefname{table}{Table}{Tables}%
  \crefname{section}{Section}{Sections}%
  \Crefname{section}{Section}{Sections}%
  \crefformat{equation}{Eq.~(#2#1#3)}%
  \Crefformat{equation}{Eq.~(#2#1#3)}%
  \crefname{proposition}{Proposition}{Propositions}%
  \Crefname{proposition}{Proposition}{Propositions}%
}

%% file: sections/introduction.tex
\section{Introduction}
\label{sec:intro}

\cyl{%
Sparse attention accelerates long-context language model decoding by reading only a subset of cached KV tokens per step, since loading the KV cache becomes the main decoding bottleneck at long context~\citep{liu2024sparsefrontier}.
Existing methods, however, commit to this subset without checking whether it captured enough attention mass: the kernel selects, attends, and emits, with no mechanism to detect or correct a bad choice.
In autoregressive decoding, even a single unchecked error compounds across the generation, as each step conditions on all previous outputs.
The compounding effect is amplified by an asymmetry in where failures occur: at tight budgets, a small minority of attention heads loses most of its softmax mass while the majority retains high coverage, yet a global budget that protects the fragile minority over-pays on the comfortable majority.
Without a per-step quality readout, the kernel cannot distinguish these two regimes or respond differently to each.
Current sparse attention is therefore open-loop: it lacks a verification signal that indicates whether the chosen subset was adequate, together with a recovery path for the cases where it was not.%
}

\cyl{%
The open-loop pattern persists because research has focused on improving the upstream selection rule rather than adding downstream verification.
Fixed-budget methods~\citep{zhang2024h2o, li2024snapkv, feng2024adakv, tang2024quest, desai2025hashattention} each propose a different scoring heuristic, and per-head budget adaptation further raises selection precision, yet every method follows the same protocol: select once per step, attend, emit, with no post-hoc quality check.
A complementary line, exemplified by Tactic~\citep{zhu2025tactic} and Twilight~\citep{lin2025twilight}, replaces the fixed budget with a target attention mass and selects tokens until that mass is reached.
\cyl{The target-mass approach partially addresses the concentrated-failure problem but not output quality, yet the mass estimate must be obtained \emph{before} the kernel runs through quantized auxiliary caches or distribution-fitting heuristics, and the resulting output is still emitted without verification.}
Concurrent methods that do close part of the loop, such as vAttention~\citep{desai2025vattention} and Vegas~\citep{yue2026specattn}, achieve verification at the cost of a sampling estimator or a full speculative-decoding pipeline.
A quality signal that requires no auxiliary infrastructure and operates inside the standard decode loop has not yet appeared.%
}

\cyl{%
The signal that the preceding discussion calls for turns out to already exist inside the sparse-attention path, hidden in numerical by-products that the decode loop computes and then discards.
The relevant quantity is the fraction of softmax mass that lands on the selected tokens, which we call \emph{coverage}.
High coverage means the sparse output closely approximates the dense one; low coverage pinpoints exactly which heads should receive additional compute.
Recent adaptive-budget methods use coverage as a \emph{pre-hoc target}: they estimate it before the kernel runs and select tokens until the estimate reaches a desired level~\citep{zhu2025tactic, lin2025twilight}.
When read \emph{after} the kernel has run rather than estimated beforehand, the same quantity costs nothing to evaluate: it reduces to a single log-sum-exp comparison between values the kernel already produces and adds under two percent to decode latency.
Post-hoc coverage is also more than a heuristic proxy: at the token level, it satisfies an exact algebraic equality with output error, and in the paged implementation a lightweight proxy preserves this relationship with empirically validated fidelity across three architectures.
Coverage therefore serves simultaneously as an acceptance criterion and as the interpolation weight for output recovery.
The signal that existing sparse-attention kernels discard after every step is, in fact, a built-in quality readout.%
}

\begin{figure}[t]
\centering
\includegraphics[width=\textwidth]{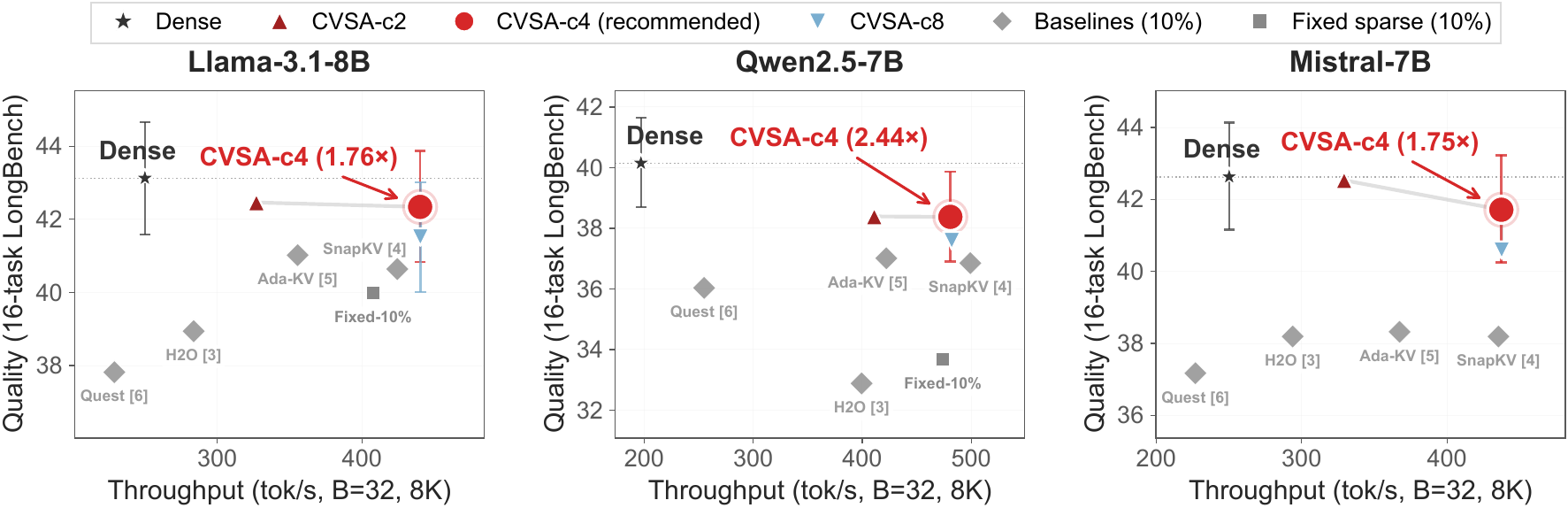}
\cyl{\caption{Quality--throughput frontier on three 7B models
(LongBench 16-task average, 8K context, batch 32).
Each marker represents one sparse-attention method at a fixed
operating point; the $x$-axis is tokens/s derived from median decode-step
latency and the $y$-axis is the benchmark score.
Open-loop baselines are grey markers.
\ssa{} combines higher quality with competitive speed; SnapKV is faster
on Qwen. Coverage checks and recovery run every $M$ steps.
\label{fig:pareto}}}
\end{figure}

\cyl{%
Building on this signal, we introduce \emph{Coverage-Verified Sparse Attention} (\ssa{}), a training-free framework that treats every sparse output as a draft and verifies it against coverage before committing.
Unlike closed-loop attempts that require a sampling estimator or a speculative-decoding pipeline, verification in \ssa{} adds no dense pass and no auxiliary model: the verifier is the coverage scalar the kernel already computes.
Only the minority of heads whose coverage falls short enter a lightweight recovery step, so amortized overhead stays close to plain sparse attention.
Across three 7B models on LongBench~\citep{bai2024longbench}, \ssa{} comes within $1.13$ points of dense attention (\cref{tab:longbench}), with median decode-step speedup reaching $6.21\times$ at 128K context.
The framework scales to 70B parameters, where it statistically matches dense attention on LongBench. On the three 7B models, it narrows the RULER gap by nearly an order of magnitude relative to the strongest prior method.
Quality is stable across a wide acceptance-threshold range, with only $0.14$ points of variation on the three-model average.
Long chain-of-thought reasoning is out of scope (\cref{app:limitations}).%
}
\cyl{%
The contributions are:
\begin{itemize}[leftmargin=*,nosep]
    \item Coverage as a free quality signal for sparse attention, grounded in an exact token-level algebraic equality between coverage and output error rather than a probabilistic bound or a sampling estimate.
    \item A verify-then-correct framework that wraps around off-the-shelf page selectors, accepts reliable sparse drafts, and applies lightweight recovery only on the heads that fail a per-head coverage check, all inside the standard decode loop with no auxiliary model.
    \item A stable operating point: quality varies by only $0.14$ points across the threshold sweep. Fixed-5\% trails dense by $5.48$ points, and \ssa{}-c4 exceeds the Fixed-12\% control by $3.57$ points at similar effective read (\cref{tab:matched-compute}); aligned controls show this margin also reflects refresh cadence and dense-layer placement (\cref{tab:aligned-controls}).
    \item LongBench quality within $1.13$ points of dense at 7B and statistically matching dense at 70B, a $0.8$-point RULER gap at 7B, and median decode-step speedup up to $6.21\times$ at 128K context.
\end{itemize}%
}

%% file: sections/preliminary.tex
\section{Preliminaries}
\label{sec:prelim}

\subsection{Attention Computation and the LSE By-Product}
\label{sec:prelim-attn}

\cyl{%
Autoregressive decoding in a decoder-only Transformer requires, at every step, a full read of the KV cache to compute one attention output.
For a query $q \in \mathbb{R}^d$ and cached keys and values $K, V \in \mathbb{R}^{n \times d}$, the dense output is $\odense = \sum_{i=1}^{n} \alpha_i v_i$ with weights $\alpha_i = \mathrm{softmax}(q^\top k_i / \sqrt{d})_i$.
Fused-attention kernels~\citep{dao2022flashattention} evaluate this sum in a single pass over the cache and, as a numerical side effect of the online softmax reduction, also produce the log-sum-exp $\mathrm{LSE} = \log \sum_{i=1}^{n} \exp(q^\top k_i / \sqrt{d})$ without another pass over the cache.
The overall bandwidth cost of the cache read scales linearly with context length and is dominated by the KV component, an imbalance that grouped-query attention (GQA)~\citep{ainslie2023gqa} magnifies by sharing each KV head across multiple query heads (ratios of $4{:}1$, $7{:}1$, and $8{:}1$ in the models evaluated in this work).
Per-head KV traffic is consequently the binding resource in decode-step latency, yet standard sparse-attention kernels allocate every head the same token budget.%
}

\subsection{Paged KV Cache and Sparse Attention}
\label{sec:prelim-paged}

\cyl{%
Production serving stacks such as vLLM~\citep{kwon2023vllm} organize the KV cache into fixed-size pages of $16$ tokens, and practical sparse-attention methods therefore operate at page granularity.
A sparse-attention kernel reads a selected subset of pages $\Sel$ and returns a draft output $\odraft$ together with a partial log-sum-exp $\mathrm{LSE}_{\Sel}$ restricted to the selected tokens.
Which pages enter $\Sel$ is decided by \emph{page scoring}: each page receives a score $\max_{j \in \text{page}}(q^\top k_j / \sqrt{d})$, and the top-$k$ pages by score form $\Sel$.
\cyl{Page scoring reads all cached keys, dominates each
sparse-attention step, and grows linearly with context length.}
The pages excluded from $\Sel$ form the complement set $\Comp$, and their contribution to the dense output is denoted $\ocomp$.
Together, $\odense$, $\odraft$, and $\ocomp$ satisfy a partition identity over $\Sel$ and $\Comp$ whose role in approximation quality analysis is formalized in \cref{sec:method-theory}.%
}

%% file: sections/method.tex
%

\section{Methodology}
\label{sec:method}



\cyl{%
\Cref{sec:coverage} formulates sparse-attention quality control as a
per-head decision problem and identifies coverage as the decision
variable; \cref{sec:method-theory} proves two exact token-level
identities whose implications extend to the recovery mechanism.
\Cref{sec:vtc} translates these results into a closed-loop pipeline,
and \cref{sec:hybrid,sec:dl} complete the system with scheduling and
dense-layer policies.%
}

\subsection{Problem Definition and Coverage}
\label{sec:coverage}

\cyl{%
Sparse attention produces a draft output $\odraft$ that may differ
from the dense output $\odense$, \cyl{and the magnitude of this difference
varies across heads, layers, and decode steps in ways that depend on
the input and are not known in advance.}
\cyl{The quality-verification problem is to determine, at runtime and
without ground-truth access, for each head independently, whether
$\odraft$ is an acceptable approximation of $\odense$.}
The central constraint is that $\odense$ itself is unavailable, as
computing it would eliminate the latency benefit that motivated sparse
attention in the first place.
A useful verifier thus must rely on quantities that the
sparse-attention path already produces as part of its normal operation.%
}

\begin{figure}[t]
\centering
\includegraphics[width=0.96\textwidth]{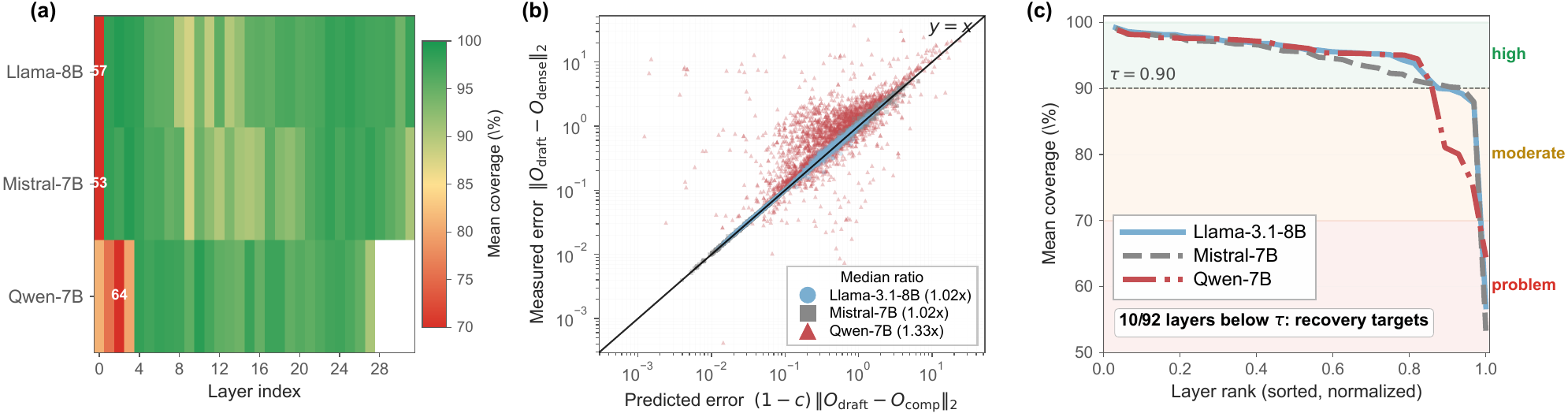}
\cyl{\caption{Empirical prerequisites for coverage-based verification,
evaluated on Llama-3.1-8B, Mistral-7B-v0.3, and Qwen2.5-7B at a
5\% page budget.
\textbf{(a)}~Per-layer attention mass retained under sparse selection
($y$-axis: fraction of dense softmax mass captured).
\cyl{Most layers retain high coverage; a small, model-dependent minority
drops sharply, confirming that sparse failures are concentrated rather
than uniform.}
\cyl{\textbf{(b)}~Measured sparse-output error ($y$-axis) vs.\ the error
predicted by \cref{thm:error} ($x$-axis), evaluated per head-step.}
The page-level proxy over-accepts on all three models, most
harmfully on Qwen through a known drift (\cref{app:bootstrap}).
\cyl{\textbf{(c)}~Sorted-rank view of per-layer coverage exposes a
thresholdable long-tail structure: only the small low-coverage
minority enters the correction path.}
\label{fig:motivation}}}
\end{figure}

\cyl{%
\cyl{The partition identity reviewed in \cref{sec:prelim-paged} points to
exactly such a quantity.}
That identity decomposes $\odense$ into contributions from $\Sel$ and
$\Comp$, and the scalar governing this decomposition is the
\emph{coverage} $c = \sum_{i \in \Sel} \alpha_i$, the fraction of
dense softmax mass captured by the sparse selection ($\alpha_i$ as
defined in \cref{sec:prelim-attn}).
\cyl{High coverage means the draft is close to $\odense$; low coverage
flags heads where the selection missed important attention mass.}
This definition is token-level; the deployed system uses a page-level
proxy validated in \cref{sec:empirical-thm}.
Whether coverage is backed by a formal quality guarantee is the
question that \cref{sec:method-theory} addresses.%
}
\cyl{%
For coverage to function as a practical quality signal, three
conditions should hold: sparse-attention failures should be
concentrated in a small minority of heads rather than spread
uniformly, the coverage value should track the output error predicted
by the token-level identity, and the low-coverage tail should be
separable enough to target with recovery.
\Cref{fig:motivation} evaluates these conditions across three 7B
model families and confirms that each is met, with the proxy
limitations on Qwen quantified in \cref{app:bootstrap}.
The concentrated-failure pattern is particularly informative: it
indicates that a verify-then-correct design can confine recovery to a
small fraction of heads at each step and keep amortized cost close to
that of plain sparse attention.%
}

\subsection{Theoretical Analysis}
\label{sec:method-theory}


\cyl{%
Coverage is not merely a useful heuristic: the two propositions below
establish it as the exact interpolation weight between the sparse
draft and the complement output, and show that it exactly determines
per-step output error.
These results hold at the token level under the softmax-attention
setup in \cref{sec:prelim-attn}, with design implications that
extend to the recovery mechanism in \cref{sec:vtc}.%
}

\begin{proposition}[Exact Decomposition]
\label{thm:decomp}
\cyl{%
For a selected subset $\Sel$ with token-level coverage
$0 < c < 1$ and complement $\Comp$,
\begin{equation}
    \odense \;=\; c \cdot \odraft \;+\; (1-c) \cdot \ocomp\,,
    \label{eq:decomp}
\end{equation}
where
$\odraft = \frac{1}{c}\sum_{i \in \Sel} \alpha_i v_i$ and
$\ocomp = \frac{1}{1{-}c}\sum_{i \in \Comp} \alpha_i v_i$.%
}
\end{proposition}


\cyl{%
The proof partitions the softmax sum into contributions from $\Sel$
and $\Comp$ and renormalizes each branch (see \cref{app:proofs}).
\Cref{eq:decomp} reveals that the dense output is a convex
combination of the draft and the complement, with coverage as the
mixing coefficient.
Recovery from an insufficient draft therefore does not require
recomputing dense attention from scratch.
It suffices to compute the complement output $\ocomp$ and blend with
the known weight $(1{-}c)$.
This converts a binary accept-or-recompute decision into an
interpolation whose cost scales with how much mass the draft missed.%
}

\begin{proposition}[Exact Error Identity]
\label{thm:error}
\cyl{%
Under the same conditions as \cref{thm:decomp},
\begin{equation}
    \|\odraft - \odense\|_2 \;=\; (1-c) \cdot \|\odraft - \ocomp\|_2
    \label{eq:error}
\end{equation}
holds as an exact equality.%
}
\end{proposition}


\cyl{%
\Cref{thm:error} follows from rearranging \cref{eq:decomp} (see
\cref{app:proofs}).
The per-step approximation error of the sparse draft is the product
of two factors: $(1{-}c)$, which measures how much attention mass the
selection missed, and $\|\odraft - \ocomp\|_2$, which measures how
different the selected and omitted content are.
When coverage is high, the prefactor $(1{-}c)$ is small and
suppresses the error regardless of the content discrepancy, which
justifies accepting the draft without further computation.
When coverage is low, the large prefactor amplifies the discrepancy,
and the identity prescribes exactly how much correction is needed
through the decomposition in \cref{eq:decomp}.
Coverage thus plays a dual role that no prior sparse-attention signal
provides: it is simultaneously the acceptance criterion (is $c$ above
$\tau$?) and the recovery weight (blend with coefficient $1{-}c$).%
}
%
\cyl{%
A direct corollary tightens the operational rule: if
$c \geq \tau$ for some threshold $\tau \in (0,1)$, then
$\|\odraft - \odense\|_2 \leq (1{-}\tau)\|\odraft - \ocomp\|_2$,
which provides multiplicative per-step error control whose tightness
is set by the threshold alone.%
}



\cyl{%
The dual role established above holds exactly at the token level.
The deployed system, however, operates at page granularity and uses
the page-level proxy introduced in \cref{sec:coverage}, which raises
the question of whether the theoretical guarantees transfer to the
practical verifier.
\Cref{app:bootstrap} provides the full validation: under token-level
coverage, predicted and measured errors match with median ratio $1.0$
on all three models; the page-level proxy over-accepts on all three
models, most harmfully on Qwen through a known drift, quantified
through false-accept and false-reject rates.
The distinction is accordingly precise: the theoretical claim is
token-level and exact, while the practical verifier is a page-level
approximation whose accuracy is measured and reported rather than
assumed.%
}
\label{sec:empirical-thm}

\subsection{Framework Overview}
\label{sec:vtc}

\begin{figure}[t]
\centering
\definecolor{cvTealFill}{HTML}{E0F2FE}
\definecolor{cvTeal}{HTML}{0E7490}
\definecolor{cvTealHot}{HTML}{5EEAD4}
\definecolor{cvGrayFill}{HTML}{F4F4F5}
\definecolor{cvGrayCell}{HTML}{E4E4E7}
\definecolor{cvGrayStroke}{HTML}{71717A}
\definecolor{cvOrangeFill}{HTML}{FED7AA}
\definecolor{cvOrange}{HTML}{C2410C}
\definecolor{cvGreenFill}{HTML}{D1FAE5}
\definecolor{cvGreen}{HTML}{059669}
\definecolor{cvRedFill}{HTML}{FEE2E2}
\definecolor{cvRed}{HTML}{DC2626}
\resizebox{\textwidth}{!}{%
\begin{tikzpicture}[
  font=\scriptsize\sffamily,
  >={Latex[length=3.6pt, width=4.2pt]},
  block/.style={rounded corners=4pt, line width=0.9pt, align=center},
  smallblock/.style={block, minimum height=0.72cm},
  hot/.style={->, line width=1.4pt, draw=cvTeal, shorten >=2pt},
  thin/.style={->, line width=0.85pt, draw=cvGrayStroke, shorten >=2pt},
  acceptline/.style={->, line width=1.1pt, draw=cvGreen, shorten >=2pt},
  recoverline/.style={->, line width=1.1pt, draw=cvRed, shorten >=2pt},
  sched/.style={densely dashed, line width=0.85pt},
  label/.style={font=\scriptsize\sffamily\bfseries, text=black!65},
  tiny/.style={font=\tiny\sffamily\bfseries, text=black!55},
  page/.style={draw=gray!55, line width=0.35pt, minimum width=0.14cm, minimum height=0.14cm, inner sep=0pt}
]
\path[fill=gray!5, draw=black!12, rounded corners=7pt] (-0.25,-0.80) rectangle (13.50,1.92);
\node[label, anchor=west] at (-0.05,1.62) {Decode step $t$: compact page-level verify-then-correct};

\path[draw=cvTeal, densely dashed, line width=0.7pt, rounded corners=5pt]
  (3.25,-0.62) rectangle (12.00,1.35);
\node[font=\scriptsize\sffamily\bfseries, text=cvTeal, anchor=west, fill=gray!5, inner sep=2pt]
  at (3.45,1.35) {CVSA};

\node[block, fill=cvGrayFill, draw=cvGrayStroke, minimum width=2.90cm, minimum height=0.78cm] (kv) at (1.65,0.38) {};
\node[tiny, fill=cvGrayFill, inner sep=1pt] at (1.65,0.85) {Paged KV cache};
\foreach \px/\fc/\dc in {0.585/cvGrayCell/gray!55, 1.165/cvTealHot/cvTeal, 1.745/cvTealHot/cvTeal, 2.655/cvOrangeFill/cvOrange} {
  \node[draw=\dc, fill=\fc, line width=0.35pt, minimum width=0.40cm, minimum height=0.40cm, inner sep=0pt] at ({\px+0.06}, 0.44) {};
  \foreach \cxi in {-1.5, -0.5, 0.5, 1.5} {
    \foreach \cyi in {-1.5, -0.5, 0.5, 1.5} {
      \node[draw=\dc, fill=\fc, line width=0.25pt, minimum width=0.10cm, minimum height=0.10cm, inner sep=0pt] at ({\px + 0.10*\cxi}, {0.38 + 0.10*\cyi}) {};
    }
  }
}
\node[font=\small\sffamily\bfseries, text=black!55] at (2.200, 0.40) {$\cdots$};
\node[draw=cvTeal, fill=cvTealHot, line width=0.4pt, minimum width=0.20cm, minimum height=0.20cm, inner sep=0pt] (legsel) at (0.82,-0.50) {};
\node[right=2pt of legsel, font=\tiny\sffamily\bfseries, text=cvTeal, inner sep=0pt] {sel.};
\node[draw=cvOrange, fill=cvOrangeFill, line width=0.4pt, minimum width=0.20cm, minimum height=0.20cm, inner sep=0pt] (legrec) at (1.50,-0.50) {};
\node[right=2pt of legrec, font=\tiny\sffamily\bfseries, text=cvOrange, inner sep=0pt] {rec.};

\node[smallblock, fill=cvTealFill, draw=cvTeal, minimum width=1.18cm, minimum height=0.88cm] (select) at (4.355,0.38) {Page sel.\\{\tiny top-$k$}\\{\tiny $\mathrm{LSE}_{\mathrm{sel}},\mathrm{LSE}_{\mathrm{all}}$}};
\node[smallblock, fill=cvTealFill, draw=cvTeal, minimum width=1.18cm, minimum height=0.88cm] (attn) at (6.38,0.38) {Sparse attn.\\{\tiny $O_{\mathrm{draft}}$}};
\node[block, fill=cvOrangeFill, draw=cvOrange, line width=1.55pt, minimum width=2.20cm, minimum height=1.30cm] (gate) at (8.50,0.45) {};
\node[font=\scriptsize\sffamily\bfseries, anchor=center] at (8.50,0.95) {$\star$ Coverage gate};
\foreach \i/\h/\fc/\dc in {%
  0/0.716/cvGreenFill/cvGreen,
  1/0.758/cvGreenFill/cvGreen,
  2/0.590/cvRedFill/cvRed,
  3/0.674/cvGreenFill/cvGreen,
  4/0.520/cvRedFill/cvRed,
  5/0.744/cvGreenFill/cvGreen%
} {
  \path[fill=\fc, draw=\dc, line width=0.3pt] ({7.80 + 0.28*\i - 0.10}, 0.10) rectangle ({7.80 + 0.28*\i + 0.10}, \h);
}
\draw[densely dashed, line width=0.55pt, draw=black!75] (7.55,0.660) -- (9.45,0.660);
\node[font=\tiny\sffamily\bfseries, anchor=south west, inner sep=1pt, text=black!75] at (9.40,0.665) {$\tau$};
\node[font=\tiny\sffamily\bfseries, anchor=center, text=black!65] at (8.50,-0.08) {$\hat c$ per KV head};

\node[smallblock, fill=cvGrayFill, draw=cvGrayStroke, minimum width=0.80cm, minimum height=0.78cm] (out) at (12.65,0.38) {$O_h$};

\node[block, fill=cvGreenFill, draw=cvGreen, minimum width=1.65cm, minimum height=0.55cm, inner sep=2pt] (accept) at (10.95,0.88) {};
\foreach \i in {0,...,3} {
  \fill[cvGreen] ({10.30 + 0.11*\i},0.88) circle (1.5pt);
}
\node[font=\scriptsize\sffamily\bfseries, anchor=west, text=cvGreen, inner sep=2pt] at (10.78,0.88) {accept};
\node[block, fill=cvRedFill, draw=cvRed, minimum width=1.65cm, minimum height=0.55cm, inner sep=2pt] (recover) at (10.95,-0.12) {};
\foreach \i in {0,...,1} {
  \fill[cvRed] ({10.30 + 0.13*\i},-0.12) circle (1.5pt);
}
\node[font=\scriptsize\sffamily\bfseries, anchor=west, text=cvRed, inner sep=2pt] at (10.58,-0.12) {recover};

\node[font=\scriptsize\sffamily\bfseries, text=cvTeal] at (4.355,-0.40) {every $M$ steps};
\node[font=\scriptsize\sffamily\bfseries, text=cvTeal] at (6.38,-0.40) {every step};
\node[font=\scriptsize\sffamily\bfseries, text=cvOrange] at (8.50,-0.40) {every $M$ steps};

\draw[hot] (kv.east) -- (select.west);
\draw[hot] (select.east) -- (attn.west);
\draw[hot] (attn.east) -- (7.40,0.38);
\draw[acceptline] (9.60,0.72) -- (accept.west);
\draw[recoverline] (9.60,0.18) -- (recover.west);
\draw[thin] (accept.east) -- (12.25,0.55);
\draw[thin] (recover.east) -- (12.25,0.21);
\end{tikzpicture}%
}
\caption{\cyl{Closed-loop decode step in \ssa{}.
Unlike open-loop sparse attention, which commits every sparse output
without feedback, \ssa{} interposes a per-head coverage gate between
the sparse kernel and the final output.
\textbf{Draft}: an off-the-shelf page selector produces $\odraft$ and
the $\mathrm{LSE}$ statistics from which coverage is computed.
\textbf{Verify}: the coverage gate compares each KV head's coverage
against $\tau$; most heads pass (green) while a small minority fails
(red).
\textbf{Correct}: accepted heads commit the draft at no additional
cost; rejected heads expand the page budget and recompute attention.
Page scoring and the coverage check run together every $M$ steps
(\cref{sec:hybrid}); sparse attention runs on every step, reusing the
selected pages in between.}}
\label{fig:cvsa-framework}
\end{figure}
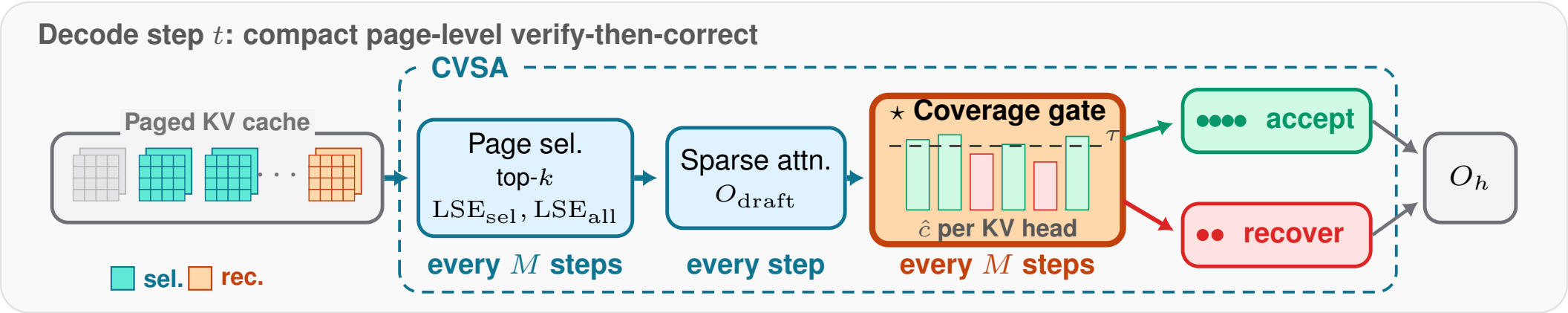

\cyl{%
The propositions establish that coverage plays a dual role in sparse
attention: it decides whether to accept a draft and prescribes the
interpolation weight for correcting one that falls short.
\ssa{} translates this dual role into a closed-loop decode procedure
with three phases: \textbf{Draft}, \textbf{Verify}, and
\textbf{Correct}, as illustrated in \cref{fig:cvsa-framework}.
In the \textbf{Draft} phase, an off-the-shelf page selector produces
a sparse output and the $\mathrm{LSE}$ values from which coverage is
estimated.
This phase is identical to the standard sparse-attention path, and
the entire \ssa{} framework is training-free: \cyl{it wraps around any
existing paged selector without modifying the model, the weights, or
the serving stack.}
In the \textbf{Verify} phase, the coverage gate evaluates each head's
coverage against a threshold $\tau$: heads above $\tau$ commit the
draft directly, and heads below $\tau$ are routed to recovery.
In the \textbf{Correct} phase, rejected heads recompute attention over
an expanded page set; \cref{app:dl-ar-ablation} compares this with the
exact $(1{-}c)$ correction of \cref{thm:decomp}.
The per-head granularity of this decision is central to efficiency:
sparse failures are localized across heads and layers, and confining
recovery to the minority that actually fails keeps amortized cost
close to that of plain sparse attention.
Two additional design choices address concerns that span multiple
steps or affect specific layers: the hybrid schedule
(\cref{sec:hybrid}) runs page scoring and the coverage check every
$M$ steps to bound how long a stale selection is reused, and the
dense-layer policy (\cref{sec:dl}) removes a structural outlier whose
coverage is persistently low regardless of input.%
}

%
%

\subsection{Hybrid Scheduling}
\label{sec:hybrid}

\cyl{%
Each coverage check compares the log-sum-exp of the selected pages'
scores with that over all pages, and both come from the page-scoring
step, which reads all cached keys and whose cost grows linearly with
context length.
\ssa{} therefore runs page scoring and the coverage check together once
every $M$ decode steps: the refreshed scores check the coverage of
the cached selection, which is kept if every head still passes;
otherwise the pages are re-selected and the heads that still fall
short enter recovery.
On the intervening steps, sparse attention reuses the selected pages
without a check.
\cyl{Reuse introduces temporal staleness, which the schedule bounds: a
head whose coverage drifts below $\tau$ within the reuse window is
caught at the next check rather than continuing on a stale selection.}
The scoring interval $M$ controls the trade-off between throughput
and selection freshness: a smaller $M$ reduces drift at the cost of
more frequent key reads, while a larger $M$ amortizes scoring more
aggressively but lets a stale selection run longer before the next
check.
\ssa{} uses $M{=}4$ as the 7B default, as this setting captures
essentially all measured throughput gains while incurring only half
the quality cost of $M{=}8$ (\cref{sec:ablations}).
This design directly addresses the error-compounding concern from the
introduction: a page selection that was adequate when chosen can go
stale as decoding progresses, and the coverage check catches this
drift by the next scoring step.%
}

%
%

\subsection{Dense Layers for Low-Sparsity Attention}
\label{sec:dl}

\cyl{%
The hybrid schedule handles temporal drift in page selections across
decode steps.
A distinct failure mode, however, is structural rather than temporal:
certain layers exhibit persistently low coverage regardless of the
input, the decode step, or the page budget.
Layer~0 is the clearest example. Early layers attend far less sparsely
than the rest of the network: \citet{tang2024quest} measure below
$10\%$ sparsity in the first two layers and exclude them from sparse
attention.
Paged sparse attention consistently fails to capture layer~0's mass at
tight budgets, and the per-layer coverage landscape across all three
model families confirms that layer~0 is a universal outlier
(\cref{app:landscape}).
Leaving layer~0 to recovery would trigger avoidable rejections at
every check, and \ssa{} therefore keeps it dense and applies
verify-then-correct to all remaining layers.
The dense mask is limited to this single layer, as other low-coverage
regions are less consistent across model families and tasks, and
expanding the mask would create an architecture-specific policy that
spends dense compute unnecessarily.
The coverage gate is designed for these variable cases: layers
that sometimes fail are caught by the per-head check, while the one
layer that always fails is handled by the fixed policy.
\Cref{sec:ablations} validates both choices empirically.%
}

%% file: sections/experiments.tex
\section{Experimental Results}
\label{sec:expts}

\subsection{Experimental Setup}
\label{sec:setup}

\cyl{%
We evaluate three 7B instruction-tuned models (GQA $4{:}1$ for
Llama-3.1-8B and Mistral-7B-v0.3, $7{:}1$ for Qwen2.5-7B) and
Llama-3.1-70B ($8{:}1$) on LongBench~\citep{bai2024longbench} (16 tasks,
50 samples; seven tasks at 70B, \cref{app:70b}) and
RULER~\citep{hsieh2024ruler} (4K--128K).
The primary baseline is Quest~\citep{tang2024quest}, a query-dependent
selector that, like \ssa{}, retains the full cache; a reproduction of
Tactic's attention-mass-target principle~\citep{zhu2025tactic} and
Twilight~\citep{lin2025twilight} are compared in
\cref{app:selector-traffic}.
H2O~\citep{zhang2024h2o}, SnapKV~\citep{li2024snapkv}, and
AdaKV~\citep{feng2024adakv} discard tokens and therefore also save
memory, which \ssa{} does not.
All baselines run at a 10\% nominal budget.
\ssa{}-c4 uses a $5\%$ base budget, $\tau{=}0.90$, $M{=}4$,
$4{\times}$ recovery budget, and dense layer~0, reading $11.7\%$ of
the KV cache (rounded: base $5.0$, recovery $3.7$, layer~0 $2.9$).
Speedups are median decode-step latency ratios on H100 (7B), H200
(long-context Qwen), and $4{\times}$H200 (70B); implementation and
statistical details are in \cref{app:bootstrap,app:reproducibility}.%
}

\subsection{Comparison with Sparse-Attention Baselines}
\label{sec:quality}

\cyl{%
We compare \ssa{}-c4 with the open-loop baselines on end-task quality
(LongBench), long-range retrieval (RULER), and decode speed, and then
examine how the default schedule transfers to 70B.%
}

\begin{figure}[t]
  \centering
  \begin{minipage}[t]{0.575\textwidth}
    \centering
    \vspace{0pt}
    \includegraphics[width=\linewidth]{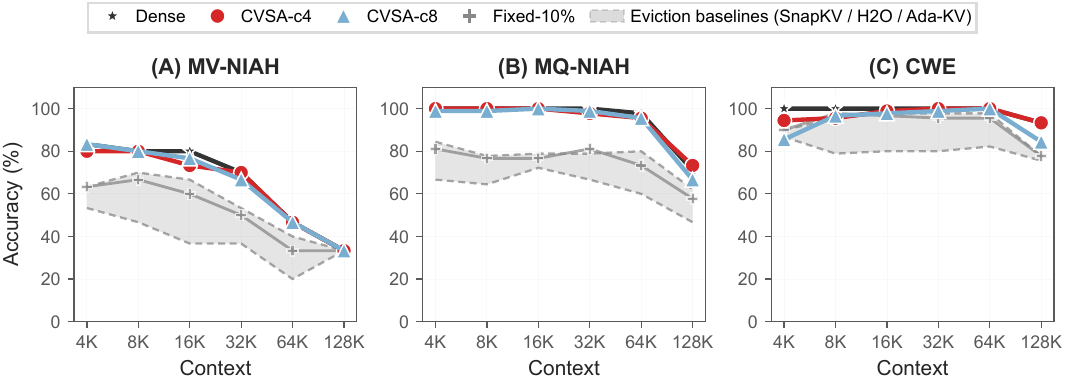}
    \captionof{figure}{\cyl{RULER retrieval stress test: accuracy on MV-NIAH, MQ-NIAH and CWE from 4K to 128K context, averaged over three 7B models. \ssa{}-c4 tracks dense; the grey band spans the open-loop baselines at $10\%$ budget (\cref{app:rulerfull}).}}
    \label{fig:ruler-main}
  \end{minipage}\hfill
  \begin{minipage}[t]{0.40\textwidth}
    \centering
    \vspace{0pt}
    \captionsetup{justification=justified,singlelinecheck=false}
    \captionof{table}{\cyl{LongBench 16-task mean (three 7B models) by design family. Baselines use a $10\%$ nominal budget; \ssa{}-c4 reads $11.7\%$ (measured).}}
    \label{tab:longbench}
    \footnotesize
    \setlength{\tabcolsep}{3.5pt}
    \renewcommand{\arraystretch}{1.0}
    \begin{tabular*}{\linewidth}{@{\extracolsep{\fill}}llcc@{}}
      \toprule
      Family & Method & Avg & Gap \\
      \midrule
      \multicolumn{2}{@{}l}{Dense} & 41.83 & --- \\
      \midrule
      \multirow{2}{*}{\emph{Query-dep.}} & \textbf{\ssa{}-c4} & \cvstabbest{40.70} & \cvstabbest{$-1.13$} \\
       & Quest & 37.01 & $-4.82$ \\
      \midrule
      \multirow{3}{*}{\emph{Eviction}} & AdaKV & \cvstabgood{38.78} & \cvstabgood{$-3.05$} \\
       & SnapKV & 38.56 & $-3.27$ \\
       & H2O & 36.67 & $-5.16$ \\
      \bottomrule
    \end{tabular*}
  \end{minipage}
\end{figure}

\begin{figure}[t]
  \centering
  \includegraphics[width=\textwidth,trim={0 4 0 0},clip]{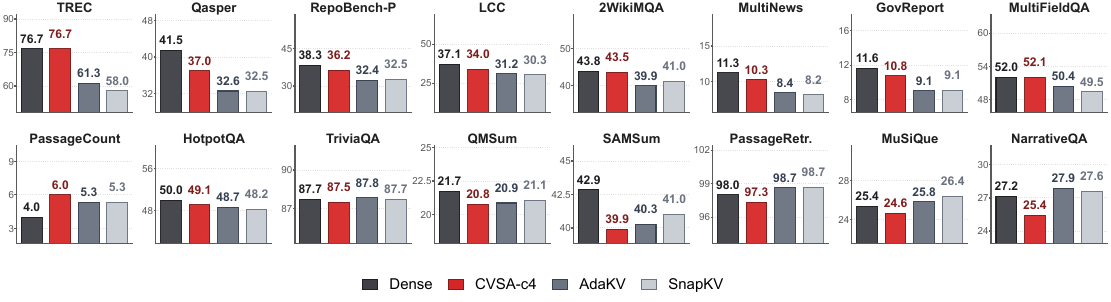}
  \caption{\cyl{Per-task LongBench scores (three-model mean), sorted by
  \ssa{}-c4's gain over the better of AdaKV and SnapKV. Extractive and
  code tasks lose most; QA and retrieval stay near dense.}}
  \label{fig:longbench-task-delta}
\end{figure}

\cyl{%
\paragraph{LongBench quality.}
Coverage-based verification closes the quality gap that upstream
selection improvements alone cannot close.
\Cref{tab:longbench} reports the three-model LongBench average:
\ssa{}-c4 is $1.13$ points below dense, whereas the strongest prior
baseline, AdaKV, is $3.05$ points below.
Unlike AdaKV, which adapts per-head budgets during selection, \ssa{}
checks coverage after sparse attention and recovers the heads that
fall short.
Paired bootstrap intervals detect small losses on all three models,
largest on Qwen (\cref{app:bootstrap}).
Raising the baseline budgets does not close the gap
(\cref{fig:budget-fairness-main}): sweeping them from 10\% to 20\%
closes at most $1.0$ point, and \ssa{}-c8 at a strictly matched
9.9\% effective budget still exceeds the strongest paged baseline by
$2.78$ points (\cref{app:isobw}).
Residual losses are concentrated (\cref{fig:longbench-task-delta}):
extractive and code-heavy tasks (Qasper $-4.48$, LCC $-3.04$, SAMSum
$-2.97$) lose more because a tight base budget can miss a critical
span before recovery has evidence to correct the selection, whereas
QA and retrieval tasks remain close to dense, indicating that
verification prevents broad quality collapse.

\paragraph{Long-range retrieval.}
\Cref{fig:ruler-main} reports RULER accuracy on the three sub-tasks
that differentiate methods (MV-NIAH, MQ-NIAH, CWE) from 4K to 128K
context.
\ssa{}-c4 tracks dense across context lengths; averaged over the six
RULER tasks at 4K--32K and three models, it loses $0.8$ points relative
to dense, compared with $8.7$ points for the strongest baseline
(\cref{app:rulerfull}).%
}

\label{sec:latency}
\cyl{%
\paragraph{Decode speed.}
These quality gains do not come at the cost of speed.
At 8K context with batch size 32 (\cref{tab:latency-8k}), a
conservative regime in which weight reads and other decode overheads
still limit the attainable speedup, \ssa{}-c4 is the fastest method on
Llama and Mistral and within $0.03\times$ of SnapKV on Qwen.
As context grows and KV reads dominate decode-step latency, the median
decode-step speedup grows monotonically, reaching $6.21\times$ at 128K
on Qwen2.5-7B (\cref{tab:long-ctx}).
This trend follows the bandwidth-roofline model (\cref{app:roofline}):
as the KV-traffic fraction $f_{\text{kv}}$ approaches unity, reducing
the number of pages read translates into a proportionally larger
end-to-end gain.
\Cref{fig:pareto} summarizes the trade-off: \ssa{}-c4 attains higher
quality than every evaluated baseline, while SnapKV is faster on Qwen.%
}

\begin{table}[t]
  \centering
  \captionsetup{justification=justified,singlelinecheck=false}
  \begin{minipage}[t]{0.60\textwidth}
    \vspace{0pt}
    \caption{\cvsaadd{Median decode-step speedup at 8K, batch 32. Eff.\ budget is measured on this benchmark ($11.7\%$
    on LongBench).}}
    \label{tab:latency-8k}
    \centering
    \small
    \setlength{\tabcolsep}{3.5pt}
    \begin{tabular*}{\linewidth}{@{\extracolsep{\fill}}lcccc@{}}
      \toprule
      Method        & Eff.\ budget & Llama & Mistral & Qwen \\
      \midrule
      \textbf{\ssa{}-c4} & \textbf{12.0\%} & $\mathbf{1.76\times}$ & $\mathbf{1.77\times}$ & $\mathbf{2.50\times}$  \\
      SnapKV        & 10\%        & $1.72\times$   & $1.74\times$     & $2.53\times$ \\
      H2O           & 10\%        & $1.72\times$   & $1.17\times$     & $2.03\times$ \\
      AdaKV         & 10\%        & $1.63\times$   & $1.47\times$     & $2.14\times$ \\
      \bottomrule
    \end{tabular*}
  \end{minipage}\hfill
  \begin{minipage}[t]{0.37\textwidth}
    \vspace{0pt}
    \caption{\cvsaadd{Median decode-step speedup on Qwen2.5-7B (H200, batch
    8); c4 reads $11.7\%$ on LongBench.}}
    \label{tab:long-ctx}
    \centering
    \small
    \setlength{\tabcolsep}{4pt}
    \renewcommand{\arraystretch}{1.25}
    \begin{tabular*}{\linewidth}{@{\extracolsep{\fill}}ccc@{}}
      \toprule
      Context & Batch & \ssa{}-c4 \\
      \midrule
      32K     & 8 & $3.07\times$ \\
      65K     & 8 & $4.44\times$ \\
      128K    & 8 & $\mathbf{6.21\times}$ \\
      \bottomrule
    \end{tabular*}
  \end{minipage}

  \vspace{6.95pt}
  \begin{minipage}[t]{0.53\textwidth}
    \vspace{0pt}
    \caption{\cvsaadd{Llama-3.1-70B, LongBench 7-task subset (samsum
    excluded, \cref{app:70b}). \ssa{}-c2 statistically matches paged dense
    (CI overlaps zero). \ssa{}-c2/c4 refresh page scores every 2/4 steps.
    AdaKV and SnapKV $\Delta$ are vs.\ KVPress dense ($54.50$).}}
    \label{tab:70b}
    \centering
    \small
    \setlength{\tabcolsep}{1.5pt}
    \renewcommand{\arraystretch}{1.3333}
    \begin{tabular*}{\linewidth}{@{\extracolsep{\fill}}lccccc@{}}
      \toprule
      Method      & Dense & \ssa{}-c2 & \ssa{}-c4 & AdaKV  & SnapKV \\
      \midrule
      Avg  & 52.02 & \textbf{52.63}  & 49.76     & 54.08  & 53.33  \\
      $\Delta$ & --- & $\mathbf{+0.60}$ & $-2.27$ & $-0.41$ & $-1.17$ \\
      \bottomrule
    \end{tabular*}
  \end{minipage}\hfill
  \begin{minipage}[t]{0.445\textwidth}
    \vspace{0pt}
    \caption{\cvsaadd{Scoring-interval sweep on the three 7B models. $M{=}4$ is the operating point, capturing essentially all of the speedup at half the quality cost of $M{=}8$. Eff.\ budget as in \cref{tab:latency-8k} ($11.7\%$ on LongBench).}}
    \label{tab:m-sweep}
    \centering
    \small
    \setlength{\tabcolsep}{3.2pt}
    \begin{tabular*}{\linewidth}{@{\extracolsep{\fill}}lccc@{}}
      \toprule
      Setting & Eff.\ budget & Speedup & Gap \\
      \midrule
      c2 & 16.1\% & $1.57\times$ & $-0.43$ \\
      \textbf{c4 default} & 12.0\% & \cvstabbest{$1.98\times$} & \cvstabbest{$-1.13$} \\
      c8 & 9.9\% & $1.99\times$ & $-2.04$ \\
      \bottomrule
    \end{tabular*}
  \end{minipage}
\end{table}
\label{sec:70b}
\cyl{%
\paragraph{Scaling to 70B.}
On the seven-task LongBench subset (\cref{tab:70b}), \ssa{}-c2
statistically matches paged dense attention ($\Delta{=}{+}0.60$, CI
$[-0.79, {+}1.95]$), whereas \ssa{}-c4 loses $2.27$ points.
This reflects the reuse window rather than the gate, which still
detects unreliable heads: with 80 layers, per-layer drift compounds
more between scoring refreshes, so deeper models need a shorter
window, and we use $M{=}2$ at 70B.
The 70B baselines run in the KVPress~\citep{jegou2024kvpress} pipeline, whose dense reference
is $2.48$ points above paged dense (\cref{app:70b}); we therefore compare
methods within each pipeline.%
}

\subsection{Ablation Studies}
\label{sec:ablations}

\begin{figure}[t]
  \centering
  \includegraphics[width=\textwidth,trim={0 6 0 20},clip]{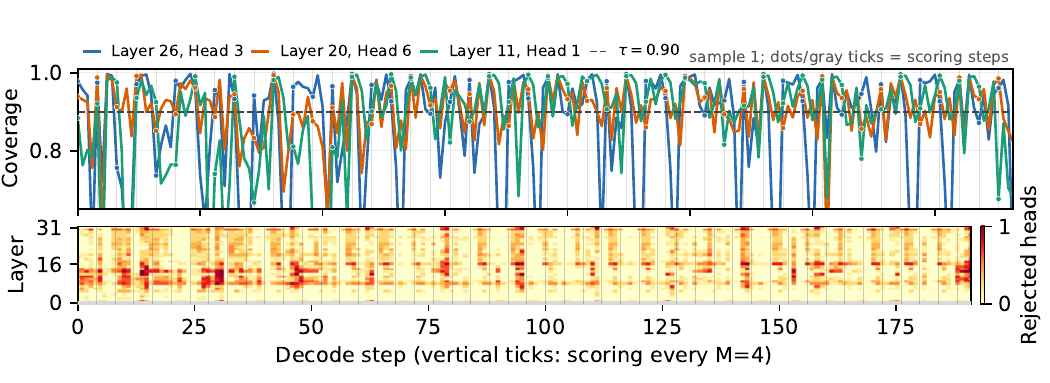}
  \caption{\cyl{Runtime coverage trace under hybrid scheduling
  ($M{=}4$, $\tau{=}0.90$, 5\% budget) on Llama-3.1-8B, HotpotQA, 8K
  context.
  \textbf{Top}: per-head coverage ($y$-axis) across 200 decode steps
  ($x$-axis) for selected KV heads; vertical marks indicate scoring
  steps where page selections are refreshed.
  Coverage jumps at scoring steps and can drift between them; heads
  found below $\tau$ (dashed line) at a check enter recovery.
  \textbf{Bottom}: per-layer rejection fraction by decode step.
  Rejection is concentrated in a few layer-step regions, which is
  consistent with the static concentrated-failure pattern in
  \cref{fig:motivation}~(a).}}
  \label{fig:hybrid-coverage-trace}
\end{figure}

\cyl{%
\Cref{fig:hybrid-coverage-trace} provides the runtime view: coverage
is refreshed at scoring steps and can drift during the reuse window. At the next check, heads below $\tau$ are re-selected and, if still short,
recovered, while the rest stay on the sparse fast path.
The bottom panel confirms that rejection is concentrated in a few
layer-step regions, consistent with the static pattern in
\cref{fig:motivation}, panel~a, and the multi-model recovery heatmaps
in \cref{fig:recovery-heatmap} confirm this across all three
architectures (mean recovered-head fraction: 21\% Llama, 23\%
Mistral, 14\% Qwen).
Recovery is therefore not a hidden dense fallback: the verification
loop spends extra compute selectively, keeping amortized
throughput close to plain sparse attention.%
}

\begin{table}[t]
  \captionsetup{justification=justified,singlelinecheck=false}
  \begin{minipage}[t]{0.555\textwidth}
    \vspace{0pt}
    \caption{\cyl{Compact verification ablation on LongBench (three 7B models; \ssa{}-c4 reads $11.7\%$ of the KV cache). \textbf{(a)}~Quality and gap to dense. \textbf{(b)}~Spread (max$-$min) of \ssa{}-c4 quality over $\tau\in[0.80,0.97]$; Avg is the spread of the three-model mean.}}
    \label{tab:tau-ablation}
    \centering
    \footnotesize
    \setlength{\tabcolsep}{3pt}
    \textbf{(a)}~Quality versus dense\\[2pt]
    \begin{tabular*}{\linewidth}{@{\extracolsep{\fill}}lccccc@{}}
      \toprule
      Configuration                & Llama  & Mistral & Qwen  & Avg    & Gap \\
      \midrule
      Dense                        & $42.57$ & $42.73$ & $40.18$ & $41.83$ & --- \\
      \midrule
      \textbf{\ssa{}-c4, $\tau{=}0.90$} & $41.87$ & $41.71$ & $38.54$ & \cvstabbest{$40.70$} & \cvstabbest{$-1.13$} \\
      \midrule
      Fixed-5\%, no gate  & $39.00$ & $37.57$ & $32.49$ & $36.35$ & \cvstabgood{$-5.48$} \\
      \bottomrule
    \end{tabular*}\par\vspace{3pt}
    \textbf{(b)}~Sensitivity to $\tau\in[0.80,0.97]$\\[2pt]
    \begin{tabular*}{\linewidth}{@{\extracolsep{\fill}}lcccc@{}}
      \toprule
      Statistic                    & Llama  & Mistral & Qwen  & Avg \\
      \midrule
      \ssa{}-c4 spread (max$-$min) & $0.38$ & $0.25$ & $0.32$ & $0.14$ \\
      \bottomrule
    \end{tabular*}
  \end{minipage}\hfill
  \begin{minipage}[t]{0.415\textwidth}
    \vspace{0pt}
    \centering
    \includegraphics[width=\linewidth]{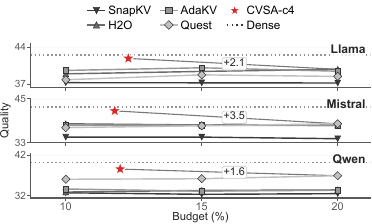}
    \vspace{-0.6em}
    \captionof{figure}{\cyl{Budget fairness on LongBench, one panel per 7B model. Curves: baselines at 10--20\% nominal budget; dotted: dense; star: \ssa{}-c4 at its $11.7\%$ effective budget; boxes: its margin over the best baseline at 20\%.}}
    \label{fig:budget-fairness-main}
  \end{minipage}
\end{table}

\cyl{%
\Cref{tab:tau-ablation} characterizes threshold sensitivity.
Moving $\tau$ across $[0.80, 0.97]$ changes the average by only
$0.14$ points. Fixed-5\% trails dense by $5.48$ points; its gap to
\ssa{}-c4 is $4.35$ points across the reported configurations.
The framework is selector-agnostic: replacing the uniform
selector with an adaptive allocator changes quality by only $0.06$
points when verification is active (\cref{app:selector-traffic}).%
}

\cyl{%
\paragraph{Verification configurations at similar effective read.}
\Cref{tab:matched-compute} compares configurations in the same paged
pipeline ($11.67\%$ effective read for \ssa{}-c4 versus $12.31\%$
for Fixed-12\% on Llama); \ssa{}-c4 leads the uniform and offline
controls by $+3.57$ and $+2.21$.
These controls also differ in refresh cadence (16 versus 4 steps);
the uniform control runs layer 0 sparse. The margins therefore do not
isolate verification. With four-step refresh and dense layer 0,
additional ungated controls reach $41.19$ (uniform) and $41.03$
(offline), versus $40.70$ for \ssa{}-c4, at somewhat higher logged
read (\cref{app:dl-ar-ablation}).
\Cref{tab:matched-compute-pertask} reports the original per-task margins.%
}

\begin{table}[t]
  \centering
  \caption{\cyl{LongBench at similar effective read: \ssa{}-c4 gains $+3.57$ over Fixed-12\% and $+2.21$ over the offline control. Controls use 16-step refresh versus four for \ssa{}-c4; Fixed-12\% also lacks dense layer 0. Aligned controls are in \cref{app:dl-ar-ablation}. \ssa{}-c4 reads $11.7\%$ of the KV cache.}}
  \label{tab:matched-compute}
  \small
  \setlength{\tabcolsep}{5pt}
  \begin{tabular*}{\textwidth}{@{\extracolsep{\fill}}lccccc@{}}
    \toprule
    Configuration (${\sim}12\%$ effective read) & Llama & Mistral & Qwen & Avg & Gap \\
    \midrule
    Dense & $42.57$ & $42.73$ & $40.18$ & $41.83$ & --- \\
    \midrule
    Uniform Fixed-12\%, no gate & $39.67$ & $38.08$ & $33.65$ & $37.13$ & $-4.70$ \\
    Offline per-(layer, head) budget, no gate & $39.98$ & $38.59$ & $36.90$ & $38.49$ & $-3.34$ \\
    \textbf{\ssa{}-c4 (coverage gate)} & $41.87$ & $41.71$ & $38.54$ & \cvstabbest{$40.70$} & \cvstabbest{$-1.13$} \\
    \bottomrule
  \end{tabular*}

  \vspace{0.6em}
  \caption{\cyl{Per-task gain of \ssa{}-c4 ($11.7\%$ read) over the Fixed-12\% control of \cref{tab:matched-compute} (three-model average). The gain is positive on all sixteen tasks and averages $+3.57$; excluding the two largest contributors (samsum, repobench-p) it is still $+3.07$.}}
  \label{tab:matched-compute-pertask}
  \footnotesize
  \setlength{\tabcolsep}{3pt}
  \begin{tabular*}{\textwidth}{@{}l@{\hspace{2\tabcolsep}}c@{\extracolsep{\fill}}l@{\hspace{2\tabcolsep}\extracolsep{0pt}}c@{\extracolsep{\fill}}l@{\hspace{2\tabcolsep}\extracolsep{0pt}}c@{\extracolsep{\fill}}l@{\hspace{2\tabcolsep}\extracolsep{0pt}}c@{}}
    \toprule
    Task & $\Delta$ & Task & $\Delta$ & Task & $\Delta$ & Task & $\Delta$ \\
    \midrule
    samsum                 & $+8.07$ & multifieldqa\_en & $+5.30$ & lcc         & $+2.23$ & musique        & $+1.59$ \\
    repobench-p            & $+6.14$ & hotpotqa         & $+4.90$ & triviaqa    & $+2.17$ & passage\_count & $+1.03$ \\
    qasper                 & $+5.90$ & 2wikimqa         & $+4.41$ & multi\_news & $+2.06$ & gov\_report    & $+0.95$ \\
    passage\_retrieval\_en & $+5.50$ & qmsum            & $+4.22$ & narrativeqa & $+2.01$ & trec           & $+0.67$ \\
    \bottomrule
  \end{tabular*}
\end{table}

\cvsadel{%
\paragraph{Scheduling, dense-layer, and recovery decisions are each
empirically grounded.}
We validate the scoring interval $M$, the dense-layer policy, and the
choice of approximate over exact recovery to confirm that each
practical decision in the framework is empirically justified.
Across $M \in \{2, 4, 8\}$, quality degrades gracefully while
throughput saturates at $M{\geq}4$, making \ssa{}-c4 the
Pareto-optimal setting for 7B models. The 70B results in
\cref{sec:70b} confirm that deeper models benefit from $M{=}2$.
Removing the layer-0 dense policy costs quality on average, with the
effect driven almost entirely by Qwen, whose layer~0 has the lowest
coverage among the three architectures (\cref{app:landscape}). The
fixed mask is accordingly a cross-model safeguard most valuable on
architectures with severe attention-sink behavior.
Exact recovery through the full complement achieves quality within a
fraction of a point of the approximate-recovery default, confirming
that expanding the page budget captures the essential mass the draft
missed at a fraction of the cost of reading the full complement.%
}

% ...

\cyl{%
\paragraph{Design decisions.}
The comparisons above concern complete operating configurations. Three practical choices within the framework also require
independent validation: the scoring interval $M$, the layer-0 dense
policy, and approximate versus exact recovery.
\Cref{tab:m-sweep} reports the scoring-interval sweep:
across $M \in \{2, 4, 8\}$, quality degrades gracefully while
throughput saturates at $M{\geq}4$, making \ssa{}-c4 the
Pareto-optimal setting for 7B models.
The 70B results confirm that deeper models benefit from $M{=}2$.
Logged telemetry agrees: within a reuse window, mean coverage drops by
$0.013$ at $M{=}2$ and $0.047$ at $M{=}8$ (two-sample Llama-3.1-8B
HotpotQA trace), the input an adaptive schedule would use
(\cref{app:limitations}).
The dense-layer policy contributes a further $+1.16$-point average
gain, driven primarily by Qwen ($+3.00$), consistent with the layer-0 analysis in \cref{sec:dl}
(per-model breakdown in \cref{app:dl-ar-ablation}).
Note that approximate recovery (AR20) matches exact recovery to within $0.05$
points on the three-model average (\cref{app:dl-ar-ablation}),
which confirms that the efficient AR path does not compromise the quality
story established by the propositions.%
}

%% file: sections/related_work.tex
\section{Related Work}
\label{sec:related}


\cyl{%
\paragraph{Open-loop sparse attention.}
Training-free sparse-attention methods reduce decode cost by choosing
a fixed subset of the KV cache and committing to it without quality
feedback.
The methods differ in how they score and select: cumulative attention
mass~\citep{zhang2024h2o}, prefill observation
windows~\citep{li2024snapkv}, per-head budget
adaptation~\citep{feng2024adakv}, query-aware page
scoring~\citep{tang2024quest}, and semantic
hashing~\citep{desai2025hashattention}.
Complementary systems address long-context inference through
attention-sink windowing~\citep{xiao2024streamingllm}, external
memory retrieval~\citep{xiao2024infllm}, head-type
separation~\citep{xiao2025duoattention}, prefill-specific sparse
patterns~\citep{jiang2024minference}, and trained sparse
architectures~\citep{yuan2025nsa}.
All share a structural limitation: once a subset is selected, the
model emits without a per-step signal that the retained attention mass
was sufficient, and without a recovery path when it was not.
\ssa{} targets this gap directly: the verification layer wraps any
open-loop selector, checks coverage post-hoc, and recovers only the
heads that fail.
The ablations in \cref{sec:ablations} examine selector compatibility,
budget, and scheduling within this framework.%
}

\cyl{%
\paragraph{Adaptive budgets and verification.}
Adaptive-budget methods move beyond fixed top-$k$ selection.
Tactic~\citep{zhu2025tactic} and Twilight~\citep{lin2025twilight}
estimate attention mass before the kernel runs and expand the
selection until the estimate is reached, reducing some fixed-budget
failures but leaving the emitted output unchecked.
Concurrent verification-oriented work closes part of the loop:
vAttention~\citep{desai2025vattention} employs a sampling estimator with
probabilistic guarantees, and
Vegas~\citep{yue2026specattn} couples sparse attention with a
speculative-decoding pipeline.
The pattern mirrors a broader trend: speculative decoding brought
verify-then-accept to token generation~\citep{leviathan2023fast,
chen2023accelerating}, and the same principle now extends to attention
computation.
The proposed \ssa{} instantiates this principle using log-sum-exp values over
page scores the selector already computes, giving a deterministic post-hoc
acceptance test and a direct recovery weight without sampling, an
auxiliary cache, or a second model.
The result is a lighter verification mechanism: no additional KV
reads for accepted heads, no stochastic estimator, and no
speculative-decoding infrastructure.
The exact algebraic equality underlying coverage (\cref{thm:error})
provides a stronger theoretical basis than the probabilistic bounds of
sampling-based approaches, while the dual role of coverage as both
acceptance criterion and recovery weight is unique to \ssa{}.%
}

\cyl{%
\paragraph{KV compression and serving systems.}
KV-cache quantization~\citep{hooper2024kvquant, liu2024kivi},
grouped-query attention~\citep{ainslie2023gqa}, and paged
management in vLLM~\citep{kwon2023vllm} reduce or reorganize
KV-cache access.
These techniques change the cost of each KV read, not the decision of
whether a sparse subset captured enough attention mass on a particular
step.
They compose naturally with \ssa{}: the verification
layer operates on top of whatever cache layout and precision the
serving stack provides, and the coverage signal remains valid
regardless of whether the underlying values are quantized or
full-precision.%
}

%% file: sections/conclusion.tex
\section{Conclusion}
\label{sec:conclusion}



\cyl{%
Coverage, the fraction of softmax mass on selected tokens, satisfies
an exact token-level equality with output error and supplies the
interpolation weight for exact recovery. \ssa{} uses a page-level
proxy for closed-loop verification with existing paged selectors,
without model modification or auxiliary infrastructure.
Quality is stable across a wide $\tau$ range. Fixed-5\% trails dense
by $5.48$ points, and \ssa{}-c4 exceeds the Fixed-12\% control by
$3.57$ points at similar effective read (\cref{tab:matched-compute});
aligned controls show this margin also reflects refresh cadence and
dense-layer placement (\cref{tab:aligned-controls}).
On LongBench at 7B scale, \ssa{} trails dense attention by $1.13$
points; on RULER (six tasks, 4K--32K), it loses $0.8$ points versus
$8.7$ for the strongest baseline.
The same framework scales to 70B with a shorter scoring interval,
where it statistically matches paged dense attention.
Measured median decode-step speedup reaches $6.21\times$ at 128K context,
Pareto-dominating every paged baseline except SnapKV on Qwen.
Residual losses concentrate on extractive and code-heavy tasks where
tight base budgets miss critical spans, which point to token-aware proxy
calibration as the most promising direction for further improvement. 
%
}

\paragraph{Limitations.}
The verifier relies on a page-level approximation of token-level
coverage that tends to over-accept; the effect is most visible on Qwen,
where false accepts concentrate in late layers; token-aware or
GQA-aware calibration could address this (\cref{app:bootstrap}).
\ssa{} targets long-context decoding with short outputs; long
chain-of-thought generation benefits from a shorter reuse window
(\cref{app:limitations}), and adaptive scheduling is a natural extension.
The speedup grows with context length and batch size and is largest on
Qwen ($6.21\times$ at 128K, $B{=}8$; \cref{app:longctx-table}).
Like Quest, \ssa{} reduces KV reads rather than KV memory, keeping the
full cache available for recovery.


%% file: appendix/algorithm.tex
\section{\ssa{} Pseudocode}
\label{app:algorithm}

\Cref{alg:cvsa} gives the deployed hybrid AR decode step for one layer.
The gate uses page-level coverage. Reuse steps perform no check, and the
base selection (not the expanded recovery set) is cached. When any head
rejects, the batched implementation computes expanded attention and masks
its outputs per KV head, including the query heads in that GQA group.

\begin{algorithm}[H]
\caption{\ssa{} hybrid AR decode step for one layer}
\label{alg:cvsa}
\begin{algorithmic}[1]
\REQUIRE query heads $Q_t$, paged KV cache $(K,V)$, optional cached set $\Sel_{t-1}$, step $t$, interval $M$, base budget $B$, threshold $\tau$, recovery budget $B_{\mathrm{AR}}$
\IF{layer is in the fixed dense set}
    \STATE return dense attention$(Q_t,K,V)$
\ENDIF
\IF{cache exists and $t \bmod M \neq 0$}
    \STATE return sparse attention on $\Sel_{t-1}$; keep the cached set
\ENDIF
\STATE score pages and compute the page-level all-context normalization
\IF{cache exists and every head's cached-set page coverage is at least $\tau$}
    \STATE return sparse attention on $\Sel_{t-1}$; keep the cached set
\ENDIF
\STATE choose a fresh base set $\Sel_t$ at budget $B$ using the page scores
\STATE run sparse attention on $\Sel_t$ to obtain $\odraft$
\STATE compute the page-level coverage $\hat c_h$ of $\Sel_t$ for each KV head $h$
\STATE set $O_h \leftarrow O_{\mathrm{draft},h}$ for all heads
\IF{any head has $\hat c_h < \tau$}
    \STATE choose an expanded set at budget $B_{\mathrm{AR}}$ and recompute attention
    \STATE replace $O_h$ with its expanded-set output only where $\hat c_h < \tau$
\ENDIF
\STATE cache the base set $\Sel_t$ and return concatenated head outputs
\end{algorithmic}
\end{algorithm}

%% file: appendix/proofs.tex
\section{Full Proofs of Propositions 1 and 2}
\label{app:proofs}

We restate and prove the two propositions used in \cref{sec:coverage}.
The setup throughout is a single decode step with query $q \in \mathbb{R}^d$, cached keys and values $K, V \in \mathbb{R}^{n \times d}$, attention weights $\alpha_i = \mathrm{softmax}(q^\top k_i / \sqrt{d}) \in (0, 1)$ with $\sum_{i=1}^{n} \alpha_i = 1$, and a partition of token indices into a selected subset $\Sel$ and its complement $\Comp = \{1, \ldots, n\} \setminus \Sel$.
Coverage is defined as $c = \sum_{i \in \Sel} \alpha_i$, so $1 - c = \sum_{i \in \Comp} \alpha_i$.

The dense, draft, and complement attention outputs are
\begin{align}
\odense  &= \sum_{i = 1}^{n} \alpha_i\, v_i,
\\
\odraft  &= \frac{1}{c} \sum_{i \in \Sel} \alpha_i\, v_i,
\qquad
\ocomp  = \frac{1}{1-c} \sum_{i \in \Comp} \alpha_i\, v_i,
\end{align}
where $\odraft$ and $\ocomp$ are the per-branch \emph{normalized} attention outputs (so each is a softmax-weighted average over its respective index subset).
We assume $0 < c < 1$ throughout; the boundary cases $c \in \{0, 1\}$ collapse trivially because one of the two branches is empty.

\subsection*{\Cref{thm:decomp} (Exact Decomposition).}
\textit{For any partition $\Sel \cup \Comp = \{1, \ldots, n\}$,}
\begin{equation*}
\odense = c \cdot \odraft + (1 - c) \cdot \ocomp.
\end{equation*}

\begin{proof}
Multiplying the definitions of $\odraft$ and $\ocomp$ by their respective normalization constants returns the unnormalized partial sums:
\begin{equation}
c \cdot \odraft = \sum_{i \in \Sel} \alpha_i v_i,
\qquad
(1 - c) \cdot \ocomp = \sum_{i \in \Comp} \alpha_i v_i.
\label{eq:partials}
\end{equation}
Summing the two equalities in \eqref{eq:partials} and using $\Sel \cup \Comp = \{1, \ldots, n\}$ with $\Sel \cap \Comp = \emptyset$,
\begin{equation*}
c \cdot \odraft + (1 - c) \cdot \ocomp
= \sum_{i \in \Sel} \alpha_i v_i + \sum_{i \in \Comp} \alpha_i v_i
= \sum_{i = 1}^{n} \alpha_i v_i
= \odense.
\qedhere
\end{equation*}
\end{proof}

The proof uses no property of the value vectors $v_i$ beyond linearity of the sum, and no property of the attention weights $\alpha_i$ beyond non-negativity and the partition $\sum_i \alpha_i = 1$.
The decomposition therefore extends without modification to GQA (group-query attention), to multi-head attention, and to any selection rule for $\Sel$ (top-$k$, sampled, learned, etc.); it is purely an algebraic identity on the softmax distribution.

\subsection*{\Cref{thm:error} (Error Equality).}
\textit{Under the same setup as \Cref{thm:decomp},}
\begin{equation*}
\| \odraft - \odense \|_2 = (1 - c) \cdot \| \odraft - \ocomp \|_2.
\end{equation*}

\begin{proof}
Subtract $\odraft$ from both sides of \Cref{thm:decomp}:
\begin{align}
\odense - \odraft
&= c \cdot \odraft + (1 - c) \cdot \ocomp - \odraft \nonumber \\
&= (c - 1) \cdot \odraft + (1 - c) \cdot \ocomp \nonumber \\
&= (1 - c) \cdot (\ocomp - \odraft). \label{eq:diff}
\end{align}
Taking the Euclidean norm of both sides of \eqref{eq:diff} and using $1 - c \geq 0$ (so $|1 - c| = 1 - c$) together with $\| \odraft - \odense \|_2 = \| \odense - \odraft \|_2$ and $\| \ocomp - \odraft \|_2 = \| \odraft - \ocomp \|_2$,
\begin{equation*}
\| \odraft - \odense \|_2
= \| \odense - \odraft \|_2
= \| (1 - c) \cdot (\ocomp - \odraft) \|_2
= (1 - c) \cdot \| \odraft - \ocomp \|_2.
\qedhere
\end{equation*}
\end{proof}

The proof yields an \emph{equality}, not an inequality: $1 - c$ is the exact multiplicative factor relating the sparse--dense disagreement $\| \odraft - \odense \|_2$ to the sparse--complement disagreement $\| \odraft - \ocomp \|_2$, with no slack.
Coverage is therefore a sufficient algebraic statistic for this identity: once $c$ and the branch disagreement $\| \odraft - \ocomp \|_2$ are known, the sparse--dense error is determined exactly.

\subsection*{Corollary (Acceptance Bound).}
\textit{If $c \geq \tau$ for some threshold $\tau \in (0, 1)$, then}
\begin{equation*}
\| \odraft - \odense \|_2 \leq (1 - \tau) \cdot \| \odraft - \ocomp \|_2.
\end{equation*}

\begin{proof}
By \Cref{thm:error} and $c \geq \tau \Rightarrow 1 - c \leq 1 - \tau$,
\begin{equation*}
\| \odraft - \odense \|_2 = (1 - c) \cdot \| \odraft - \ocomp \|_2 \leq (1 - \tau) \cdot \| \odraft - \ocomp \|_2.
\qedhere
\end{equation*}
\end{proof}

The corollary is the formal version of the token-level per-step acceptance guarantee: whenever an exact coverage verifier accepts ($c \geq \tau$), the sparse output is within $(1 - \tau) \cdot \| \odraft - \ocomp \|_2$ of dense.
A simpler $D_V$-based form replaces $\| \odraft - \ocomp \|_2$ with the value diameter $D_V = \max_{i, j} \| v_i - v_j \|_2$, yielding $\| \odraft - \odense \|_2 \leq (1 - \tau) \cdot D_V$.
The $D_V$ form is looser by roughly $14\times$ in our measurements; the equality of \Cref{thm:error} is the primary theoretical and empirical result.

\subsection*{Implementation note: $c^\star$ vs.\ $\hat c$.}
The proofs above use the algebraic coverage $c = \sum_{i \in \Sel} \alpha_i$ defined over the full token softmax distribution.
In our paged-decode implementation we never materialize the full token-level attention scores; instead, we compute a page-max proxy $\hat c = \exp(\mathrm{LSE}^{\text{page}}_{\Sel} - \mathrm{LSE}^{\text{page}}_{\text{all}})$ that is a free by-product of the page-scoring kernel.
\Cref{thm:decomp} and \cref{thm:error} are stated for the true coverage $c$ (or equivalently, the token-level $c^\star = \exp(\mathrm{LSE}^{\text{tok}}_{\Sel} - \mathrm{LSE}^{\text{tok}}_{\text{all}})$); the equality is validated under $c^\star$ on the three 7B models (4:1 and 7:1 GQA), while $\hat c$ is evaluated as the runtime proxy used by the paged system (\cref{fig:theory-validation} and \cref{sec:empirical-thm}).

%% file: appendix/bootstrap.tex
\section{Bootstrap Confidence Intervals and Theory Validation}
\label{app:bootstrap}

\begin{figure}[t]
\centering
\includegraphics[width=\textwidth]{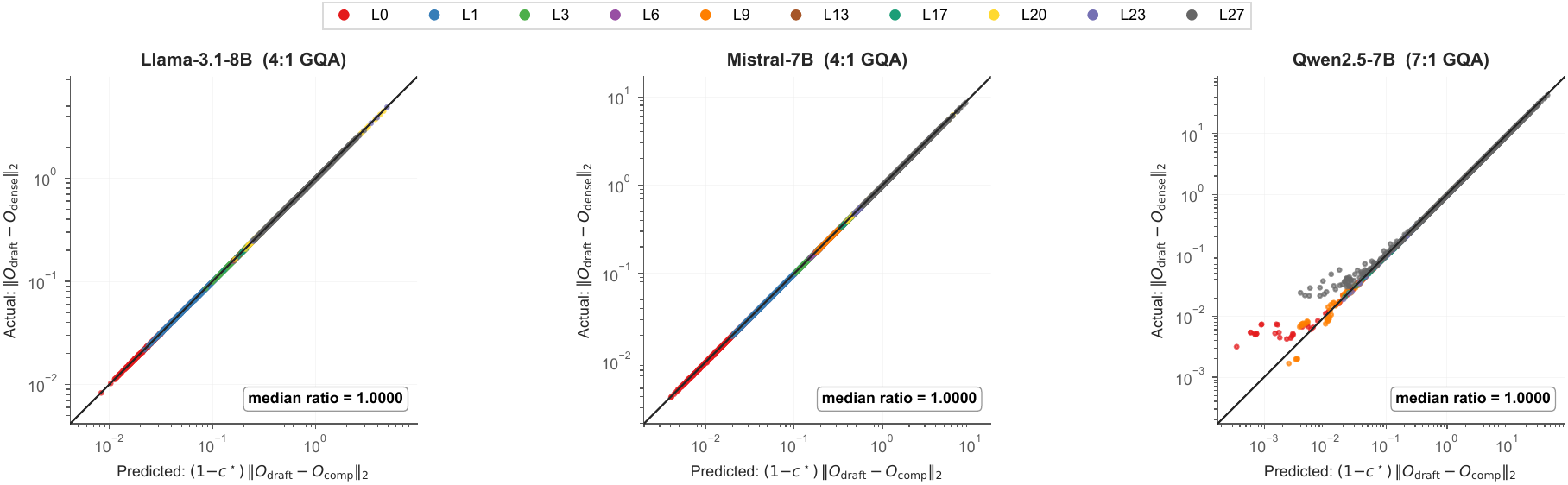}
\caption{Empirical validation of \Cref{thm:error} under token-level coverage $c^\star$, on three 7B architectures (12 prompts $\times$ 10 layers $\times$ 4 budgets $\times$ 3 thresholds). Predicted error $(1{-}c^\star)\|\odraft - \ocomp\|_2$ matches measured error $\|\odraft - \odense\|_2$ on the $y{=}x$ line per head-step; median ratio is $1.0000$ on all three models, color-coded by layer. Qwen's tail scatter at small predicted error reflects bfloat16 numerical noise where $(1{-}c^\star)$ approaches zero; the equality holds exactly in the median.}
\label{fig:theory-validation}
\end{figure}

\paragraph{\Cref{thm:error} validation under $c^\star$.}
\cref{fig:theory-validation} validates \Cref{thm:error} on $132{,}480$ logged head-steps across three 7B architectures (Llama-3.1-8B, 4:1 GQA; Mistral-7B-v0.3, 4:1 GQA; Qwen2.5-7B, 7:1 GQA), using the token-level coverage $c^\star = \exp(\mathrm{LSE}^{\text{tok}}_{\Sel} - \mathrm{LSE}^{\text{tok}}_{\text{all}})$ rather than the page-max proxy $\hat c$ used in code.
The median ratio of measured error to predicted error is exactly $1.0000$ on all three models, confirming the equality holds empirically across GQA factors.
The visible scatter on the Qwen panel is concentrated at small predicted error where $(1{-}c^\star) \to 0$ amplifies bfloat16 numerical noise; the median is unaffected.

\paragraph{$D_V$ corollary.}
The $D_V$ corollary $\|\odraft - \odense\| \leq (1{-}c) \cdot D_V$ holds with zero violations on Llama and Mistral under both $\hat c$ and $c^\star$ ($46{,}080$ cells per model).
On Qwen ($40{,}320$ cells), $\hat c$ gives $3.30\%$ violations concentrated in late layers (L17 $7.2\%$, L27 $21.7\%$) where $\hat c$ saturates near $1$ and the proxy bound $(1{-}\hat c) D_V$ becomes spuriously tight; switching to $c^\star$ reduces violations to $0.60\%$.
The residual is attributable to the sampled-$D_V$ approximation (the implementation samples up to 200 value vectors per head to estimate the diameter, which can underestimate the true geometric diameter for heavy-tailed value distributions), not to corollary failure.

\paragraph{Proxy accept/reject calibration.}
\cref{tab:proxy-calibration} compares the runtime page-level accept decision against the token-level accept decision on the same theorem-validation grid, restricted to the paper's main threshold $\tau=0.90$ and 5\% base budget.
This is a stress diagnostic rather than the production accept-rate distribution: it uses 12 HotpotQA prompts, next-token queries, and 10 selected layers per model.
Within this grid, Llama and Mistral have no accept cells under either proxy or token coverage at $\tau=0.90$, so the proxy makes no false accepts.
Qwen is the important caveat: the page-max proxy accepts 413/3360 cells, while token coverage accepts 146/3360; 315 cells are false accepts (9.4\% of all cells, 76.3\% of proxy accepts) and 48 are false rejects (1.4\% of all cells).
The false accepts concentrate in late layers (especially L17/L23/L27), consistent with the Qwen proxy drift discussed in \cref{sec:empirical-thm}.

\begin{table}[t]
  \centering
  \small
  \caption{Proxy calibration at $\tau=0.90$, 5\% budget. ``Proxy accept'' uses the page-level $\hat c$ used in code; ``Token accept'' uses token-level $c^\star$. False accept/reject rates are fractions of all cells.}
  \label{tab:proxy-calibration}
  \begin{tabular}{lcccccc}
    \toprule
    Model & Cells & Proxy accept & Token accept & False accept & False reject & Agreement \\
    \midrule
    Llama-3.1-8B & 3840 & 0.0\% & 0.0\% & 0.0\% & 0.0\% & 100.0\% \\
    Mistral-7B-v0.3 & 3840 & 0.0\% & 0.0\% & 0.0\% & 0.0\% & 100.0\% \\
    Qwen2.5-7B & 3360 & 12.3\% & 4.3\% & 9.4\% & 1.4\% & 89.2\% \\
    \bottomrule
  \end{tabular}
\end{table}

\paragraph{Production-distribution confusion matrix.}
The grid above is a controlled stress diagnostic; \cref{tab:fa-production}
reports the same accept/reject confusion matrix at the operating point
($\tau{=}0.90$, 5\% base) over the production distribution, a probe
of $161{,}280$--$280{,}640$ head-steps per model, $48$--$73\times$ the
size of the stress grid.
The proxy is optimistic on every model: false accepts are $31\%$,
$33\%$, and $39\%$ of proxy accepts on Llama, Mistral, and Qwen.
What differs across models is the consequence rather than the rate:
the median output error of a false accept is $0.19$, $0.27$, and
$0.61$ (absolute), and the catastrophic fraction (true coverage
$c^\star < 0.5$) is $4\%$, $4\%$, and $14\%$.
On Qwen these false accepts carry $6.4\times$ the output error of a
true accept and concentrate late in the network, with layers
$\geq 17$ holding $51\%$ of them, which is the mechanism behind the
$-1.64$ Qwen gap in \cref{tab:tau-ablation}.
False rejects are the benign direction: $8.4\%$, $10.2\%$, and
$15.8\%$ of proxy rejects (so $84$--$92\%$ of rejects are justified),
with a median true coverage of $0.94$--$0.95$; their only cost is
recovery pages that the $11.7\%$ effective-read accounting already
includes.

\begin{table}[t]
  \centering
  \small
  \caption{Accept/reject confusion matrix over the production distribution ($161{,}280$--$280{,}640$ head-steps per model, $48$--$73\times$ the stress grid of \cref{tab:proxy-calibration}). The proxy over-accepts on all three models, but the harmful tail (high-error and catastrophic false accepts) is confined to Qwen and to late layers ($\geq 17$ hold $51\%$), consistent with 7:1 GQA pooling concealing the worst group member. False rejects are quality-harmless and their cost is already counted in the $11.7\%$ effective read.}
  \label{tab:fa-production}
  \begin{tabular}{lccc}
    \toprule
    Production probe ($\tau{=}0.90$, 5\% base) & Llama-3.1-8B & Mistral-7B-v0.3 & Qwen2.5-7B \\
    \midrule
    False accepts (\% of proxy accepts) & $31\%$ & $33\%$ & $39\%$ \\
    FA output-error median (absolute) & $0.19$ & $0.27$ & $0.61$ \\
    Catastrophic FA ($c^\star < 0.5$) & $4\%$ & $4\%$ & $14\%$ \\
    \midrule
    False rejects (\% of proxy rejects) & $8.4\%$ & $10.2\%$ & $15.8\%$ \\
    \bottomrule
  \end{tabular}
\end{table}

\paragraph{Aggregate and per-model bootstrap intervals.}
\cref{tab:bootstrap-vsa-dense} reports the bootstrap intervals used in the main text.
The three-model main-table scores use 50 samples per task; the per-model
intervals use the separate 100-sample evaluation. These bases are labeled
explicitly. Marginal score intervals overlap, while paired $\Delta$ intervals
exclude zero for all three models. The mean losses remain below two points.

\begin{table}[t]
  \centering
  \small
  \caption{LongBench point estimates and bootstrap intervals for \ssa{}-c4 vs.\ dense (16 tasks, greedy decoding; $n$ samples per task). Means are task-averaged sample scores. For the $n=100$ rows, 10{,}000 resamples (NumPy default RNG, seed 42) draw paired example indices within each task, with tasks fixed. The 95\% percentile intervals for $\Delta$ use the same draw for both methods; all three exclude zero. The $n=50$ aggregate retains the main-table point estimates, without attaching an interval from the $n=100$ experiment. \ssa{}-c4 reads $11.7\%$ of the KV cache.}
  \label{tab:bootstrap-vsa-dense}
  \begin{tabular}{lcccc}
    \toprule
    Model & Dense mean & \ssa{}-c4 mean & 95\% CI(\ssa{}) & 95\% CI($\Delta$) \\
    \midrule
    3-model avg ($n=50$) & 41.83 & 40.70 & --- & --- \\
    Llama-3.1-8B ($n=100$) & 43.13 & 42.34 & [40.83, 43.87] & [$-1.27$, $-0.32$] \\
    Mistral-7B-v0.3 ($n=100$) & 42.63 & 41.71 & [40.25, 43.23] & [$-1.44$, $-0.38$] \\
    Qwen2.5-7B ($n=100$) & 40.15 & 38.38 & [36.91, 39.87] & [$-2.32$, $-1.24$] \\
    \bottomrule
  \end{tabular}
\end{table}

\paragraph{Paired comparison against KVPress baselines.}
\cref{tab:paired-kvpress-ci} reports paired bootstrap CIs for \ssa{}-c4 against the canonical KVPress SnapKV and AdaKV implementations on the shared 50 samples/task prefix.
Five of six comparisons exclude zero; the remaining comparison (Llama vs.\ AdaKV) is positive but borderline.

\begin{table}[t]
  \centering
  \small
  \caption{Paired bootstrap CIs for \ssa{}-c4 minus KVPress baselines on LongBench. Positive values favor \ssa{}; \ssa{}-c4 reads $11.7\%$ of the KV cache.}
  \label{tab:paired-kvpress-ci}
  \begin{tabular}{lcc}
    \toprule
    Model & \ssa{}-c4 $-$ SnapKV & \ssa{}-c4 $-$ AdaKV \\
    \midrule
    Llama-3.1-8B & $+1.23$ [+$0.25$, +$2.26$] & $+0.85$ [$-0.15$, +$1.88$] \\
    Mistral-7B-v0.3 & $+3.52$ [+$2.35$, +$4.71$] & $+3.39$ [+$2.27$, +$4.54$] \\
    Qwen2.5-7B & $+1.69$ [+$0.56$, +$2.86$] & $+1.53$ [+$0.44$, +$2.65$] \\
    \bottomrule
  \end{tabular}
\end{table}

\paragraph{Cross-pipeline dense matching.}
SnapKV and AdaKV quality numbers use KVPress in HuggingFace SDPA, while \ssa{} uses the paged-decode pipeline.
\cref{tab:dense-match} reports the dense reference in both pipelines.
Llama matches within 0.2 pt; Mistral and Qwen show larger but still sub-2-pt drift, so we use paired within-pipeline comparisons where possible and avoid ranking 70B methods by absolute cross-pipeline score.

\begin{table}[t]
  \centering
  \small
  \caption{Dense-reference alignment between the paged-decode pipeline and KVPress/HuggingFace SDPA. \ssa{}-c4 reads $11.7\%$ of the KV cache.}
  \label{tab:dense-match}
  \begin{tabular}{lccc}
    \toprule
    Model & Paged dense & KVPress dense & KVPress $-$ paged \\
    \midrule
    Llama-3.1-8B & 42.57 & 42.70 & +0.13 \\
    Mistral-7B-v0.3 & 42.73 & 41.03 & $-1.70$ \\
    Qwen2.5-7B & 40.18 & 39.27 & $-0.91$ \\
    \bottomrule
  \end{tabular}
\end{table}

\paragraph{Single-pipeline rerun of all baselines.}
To remove the mixed-pipeline confound entirely, we re-ran SnapKV and
AdaKV inside the same paged-decode pipeline as every other method, so
that all six configurations share one dense reference
(\cref{tab:unified-paged}).
Under this single yardstick \ssa{}-c4's margin over the strongest
baseline grows from $+1.92$ to $+3.70$ points, which indicates that
the mixed-pipeline table in the main text is conservative for \ssa{}
rather than favourable to it.
The margins for SnapKV and AdaKV widen because their token-level and
prefill-observation mechanisms lose precision when adapted to
$16$-token pages, whereas coverage is page-native: it is read from the
page scores that page selection already computes.
We therefore keep the mixed-pipeline numbers in the main text, where
each baseline runs in its authors' intended setting, and report this
rerun as the stricter check.

\begin{table}[t]
  \centering
  \small
  \setlength{\tabcolsep}{5pt}
  \caption{Unified single-pipeline LongBench comparison (16 tasks, 50 samples per task, greedy decoding). All six methods run in the same paged-decode pipeline against one dense reference, removing the mixed-pipeline confound of \cref{tab:dense-match}. SnapKV and AdaKV are paged-decode adaptations of their released implementations and lose accuracy relative to their native token-level setting, which is why their margins widen here; the main-text comparison, where each baseline runs in its intended framework, is the conservative one for \ssa{}. \ssa{}-c4 reads $11.7\%$ of the KV cache.}
  \label{tab:unified-paged}
  \begin{tabular}{lccccc}
    \toprule
    Method (all paged-decode) & Llama & Mistral & Qwen & Avg & Gap \\
    \midrule
    Dense                         & $42.57$ & $42.73$ & $40.18$ & $41.83$ & --- \\
    \textbf{\ssa{}-c4 (5\% base)} & $41.87$ & $41.71$ & $38.54$ & $\mathbf{40.70}$ & $\mathbf{-1.13}$ \\
    AdaKV (10\%)                  & $39.59$ & $37.82$ & $33.63$ & $37.01$ & $-4.82$ \\
    Quest (10\%)                  & $37.82$ & $37.17$ & $36.03$ & $37.01$ & $-4.82$ \\
    H2O (10\%)                    & $38.94$ & $38.19$ & $32.88$ & $36.67$ & $-5.16$ \\
    SnapKV (10\%)                 & $35.96$ & $33.09$ & $31.99$ & $33.68$ & $-8.15$ \\
    \bottomrule
  \end{tabular}
\end{table}

\paragraph{Stability over long decode.}
The coverage traces in the main text cover a few hundred decode steps.
To check whether the selection drifts over much longer generations, we
ran a $32{,}768$-step trace on Llama-3.1-8B with the KV cache growing
from 8K to 41K tokens, in which the model decodes densely and the c4
schedule runs alongside it, logging coverage and recovery decisions.
Across the entire second half of the trace (decode depths 16K to 32K),
per-head coverage holds between $0.935$ and $0.937$ per $4{,}096$-step
bin and the recovery rate holds between $18.2\%$ and $18.9\%$, with
the within-window verification signature intact at the final step.
Selection quality therefore settles at a stable operating point rather
than degrading with decode depth, because the selection and the
certificate are recomputed against the current query every $M$ steps.
Since the trace decodes densely, it measures selection quality, not
whether sparse-output errors compound.

\paragraph{Stochastic decoding.}
All LongBench and RULER numbers in the paper use greedy decoding for
reproducibility, which is standard in this family and applied
identically to every method.
The verification loop itself does not depend on that choice: coverage
is computed from the current query's page scores and log-sum-exp
before any token is drawn, so it is a property of the attention
distribution rather than of the sampling rule.
A direct check confirms this.
Under seed-paired nucleus sampling on Llama-3.1-8B (temperature $0.6$, top-$p$ $0.9$)
over eight tasks that include the two on which \ssa{}-c4 performs
weakest, its gap to dense is $-0.63$, against $-1.31$ for greedy
decoding on the same tasks: the gap does not widen under stochastic
decoding.

%% file: appendix/longbench.tex
\section{LongBench 16-task Per-Task Breakdown}
\label{app:pertask}
\cvsadel{\todo{Per-task numbers for all three 7B models; per-task $\Delta$ vs.\ dense.}}


\cyl{%
\Cref{tab:longbench-pertask-delta} expands the aggregate
LongBench result reported in the main text into per-task scores,
sorted by three-model mean gap so that the failure modes appear
first.
The pattern is concentrated rather than uniform: the largest residual
losses occur on Qasper ($-4.48$), LCC ($-3.04$), SAMSum ($-2.97$),
and RepoBench-P ($-2.12$), while retrieval and classification tasks
are within noise of dense.
These four tasks share a structural property: the information the
model must reproduce is a short, position-sensitive span (a specific
answer passage, a code continuation, or a dialogue turn) embedded in
a long context.
A 5\% base budget may exclude the page containing that span, and
the coverage verifier can only flag the miss after the pages are
selected; if the critical span happens to contribute little to the
overall softmax mass, coverage may not drop below $\tau$ even though
the output is degraded.
This failure mode points to a concrete future direction: token-aware
or query-conditioned proxy calibration that would tighten the
coverage signal on position-sensitive content.
By contrast, broader QA and retrieval tasks tolerate the same sparse
draft because the answer signal is distributed across multiple pages,
making it unlikely that all relevant mass is excluded simultaneously.%
}

\begin{table}[t]
  \centering
  \caption{\cvsaadd{LongBench per-task disclosure for the three 7B models.
  Dense and \ssa{} columns are three-model mean scores; $\Delta$ is
  \ssa{}-c4 minus dense.  The last three columns show the per-model
  $\Delta$, exposing where the aggregate loss comes from. \ssa{}-c4 reads $11.7\%$ of the KV cache.}}
  \label{tab:longbench-pertask-delta}
  \begingroup
  \color{CVSAAdd}
  \scriptsize
  \setlength{\tabcolsep}{3.0pt}
  \renewcommand{\arraystretch}{0.92}
  \begin{tabular*}{\textwidth}{@{\extracolsep{\fill}}lrrrrrr@{}}
    \toprule
    Task & Dense & \ssa{} & Mean $\Delta$ & Llama $\Delta$ & Mistral $\Delta$ & Qwen $\Delta$ \\
    \midrule
    Qasper & 41.47 & 36.99 & $-4.48$ & $-4.69$ & $-5.01$ & $-3.73$ \\
    LCC & 37.08 & 34.04 & $-3.04$ & $-3.44$ & $-4.82$ & $-0.86$ \\
    SAMSum & 42.87 & 39.90 & $-2.97$ & $-1.67$ & $-2.24$ & $-5.00$ \\
    RepoBench-P & 38.33 & 36.21 & $-2.12$ & $-1.58$ & $-3.28$ & $-1.50$ \\
    NarrativeQA & 27.17 & 25.44 & $-1.72$ & $-1.65$ & $-2.52$ & $-1.00$ \\
    MultiNews & 11.29 & 10.31 & $-0.98$ & $-0.30$ & $-1.55$ & $-1.08$ \\
    QMSum & 21.74 & 20.77 & $-0.97$ & $+0.01$ & $-0.03$ & $-2.89$ \\
    HotpotQA & 50.02 & 49.11 & $-0.91$ & $-0.37$ & $-0.94$ & $-1.42$ \\
    GovReport & 11.64 & 10.75 & $-0.89$ & $+0.00$ & $-0.79$ & $-1.87$ \\
    MuSiQue & 25.37 & 24.64 & $-0.73$ & $-0.55$ & $+0.09$ & $-1.72$ \\
    Passage Retr. & 98.00 & 97.33 & $-0.67$ & $+0.00$ & $+2.00$ & $-4.00$ \\
    2WikiMQA & 43.85 & 43.54 & $-0.31$ & $-2.14$ & $+1.42$ & $-0.20$ \\
    TriviaQA & 87.71 & 87.48 & $-0.23$ & $-1.24$ & $+1.20$ & $-0.67$ \\
    TREC & 76.67 & 76.67 & $+0.00$ & $+0.00$ & $+0.00$ & $+0.00$ \\
    MultiFieldQA & 52.05 & 52.05 & $+0.01$ & $+2.29$ & $-1.87$ & $-0.39$ \\
    Passage Count & 4.00 & 6.00 & $+2.00$ & $+4.00$ & $+2.00$ & $+0.00$ \\
    \bottomrule
  \end{tabular*}
  \endgroup
\end{table}

%% file: appendix/m_sweep.tex
\section[Check-Interval M Sweep]{Check-Interval $M$ Sweep (Quality, Latency, Effective Budget)}
\label{app:m-sweep}

\cref{tab:m-sweep} (\cref{sec:hybrid}) summarizes the three-model average $M$-sweep that motivates \ssa{}-c4 as the 7B default.
This appendix expands the per-model quality breakdown and explains why $M{=}4$ is the Pareto choice.

\paragraph{Why Pareto-optimal at $M{=}4$.}
Three forces interact as $M$ grows:
\textbf{(i) Effective budget falls} from $16.1\%$ at $M{=}2$ (frequent re-scoring + AR) to $9.9\%$ at $M{=}8$ (rare re-scoring, accept-cached selection),
\textbf{(ii) Page-scoring throughput cost amortizes}: scoring is the expensive non-attention step in any paged sparse-attention method, and beyond $M{=}4$ the per-step amortized cost is already negligible; the speedup curve flattens between $M{=}4$ ($1.98\times$) and $M{=}8$ ($1.99\times$),
\textbf{(iii) Quality degrades super-linearly} once the cached selection drifts: $\Delta$ moves from $-0.43$ at $M{=}2$ to $-1.13$ at $M{=}4$ (a $0.70$ pt jump) to $-2.04$ at $M{=}8$ (a $0.91$ pt jump).
$M{=}4$ is the elbow: it captures essentially all of the speedup ($1.98\times$ vs.\ $1.99\times$ at $M{=}8$) while paying only half the quality cost of $M{=}8$.

\paragraph{Per-model quality breakdown.}
\cref{tab:m-sweep-per-model} reports $M$-sweep quality on each of the three 7B models (LongBench 16-task average, $5\%$ budget, AR$=20\%$, dL0).
\ssa{}-c2 stays close to dense on Llama and Mistral and within $1$ pt on Qwen; \ssa{}-c4 incurs a small additional drop on each model; \ssa{}-c8 drops more on Qwen (the 7:1 GQA model) than on the others, mirroring the page-max proxy degradation discussed in \cref{sec:empirical-thm}.
On every model, c4 stays close to c2 quality at $\sim$25\% lower effective budget, so the three-model Pareto observation also holds per model.

\begin{table}[t]
  \centering
  \small
  \caption{$M$-sweep per-model quality at $5\%$ budget, AR$=20\%$, dL0. The c8 row uses the completed LongBench-16 runs for all three models. The three-model average ($-2.04$) matches \cref{tab:m-sweep}. On LongBench c4 reads $11.7\%$ of the KV cache.}
  \label{tab:m-sweep-per-model}
  \begin{tabular}{lccccc}
    \toprule
    Configuration              & Llama  & Mistral & Qwen   & 3-model avg & $\Delta$ vs.\ dense \\
    \midrule
    Dense                      & $42.57$ & $42.73$ & $40.18$ & $41.83$    & ---     \\
    \midrule
    \ssa{}-c1                  & $42.67$ & $42.47$ & $39.45$ & $41.53$    & $-0.30$ \\
    \ssa{}-c2                  & $42.49$ & $42.53$ & $39.19$ & $41.40$    & $-0.43$ \\
    \textbf{\ssa{}-c4 (default)} & $\mathbf{41.87}$ & $\mathbf{41.71}$ & $\mathbf{38.54}$ & $\mathbf{40.70}$ & $\mathbf{-1.13}$ \\
    \ssa{}-c8                  & $41.18$ & $40.59$ & $37.59$ & $39.79$    & $-2.04$ \\
    \bottomrule
  \end{tabular}
\end{table}

\paragraph{Latency and effective budget per $M$.}
The latency and effective-budget figures in \cref{tab:m-sweep} (\cref{sec:hybrid}) are three-model averages on H100 80GB at 8K context, batch size 32, measured in our paged-decode pipeline.
At 32K--128K, \ssa{}-c8 stays within $0.03\times$ of \ssa{}-c4's median decode-step speedup on all three models, so the longer interval buys almost no speed at long context.

\paragraph{The $M{=}1$ endpoint.}
Refreshing the page selection at every step ($M{=}1$) is the
quality-premium rather than the throughput configuration.
On the LongBench 16-task average it recovers most of the remaining gap
to dense, reaching $\Delta = -0.30$ across the three models against $-1.13$ for
the deployed $M{=}4$, with Llama at $+0.10$, within noise of dense
(\cref{tab:m-sweep-per-model}).
Its cost is speed: at 8K context and $B{=}32$ it runs at $0.93\times$
dense decode, so per-step scoring cancels the sparse-attention
speedup; memory is unchanged from dense since the full KV cache is
retained regardless of $M$.
$M{=}1$ is thus the setting to select when exact per-step coverage is
required at a known throughput cost, and it is the configuration whose
reasoning-length behavior is analyzed in \cref{app:limitations}.

\paragraph{Recommendation.}
\ssa{}-c4 ($M{=}4$) is the recommended 7B default; \ssa{}-c2 ($M{=}2$) is the recommended setting at $\geq 70$B, where additional layers amplify error accumulation across the longer reuse window (\cref{app:70b}).

%% file: appendix/dense_recovery_ablation.tex
\section{Dense-Layer and Recovery Ablations}
\label{app:dl-ar-ablation}

This appendix expands the ablation analysis in \cref{sec:ablations}, separating two possible confounds in the main \ssa{} configuration: the fixed layer-0 dense policy (dL0) and approximate recovery (AR20).
All results use LongBench 16-task averages with 50 samples per task on the three 7B models.

\paragraph{Dense-layer policy.}
\cref{tab:dl-ablation-full} compares fixed 5\% sparsity, \ssa{} without dL0, and the default \ssa{} configuration with dL0.
The fixed-5\% no-verification baseline trails dense by $5.48$ pt on the three-model average.
The verified AR20 configuration without dL0 improves the average from $36.35$ to $39.54$, recovering $3.19$ pt.
Adding dL0 improves the average further to $40.70$, a $+1.16$ pt gain over no-dL0.
The dL0 effect is not uniform: Llama changes by $-0.11$ pt, Mistral gains $+0.60$ pt, and Qwen gains $+3.00$ pt.
This supports the framing used in \cref{sec:dl}: dL0 is a simple cross-model safeguard for the universal low-coverage layer, not a model-specific dense mask, but it is especially important for Qwen.

\begin{table}[t]
  \centering
  \small
  \caption{Dense-layer ablation. The verified AR configuration improves on Fixed-5\%; its refresh cadence also differs. Within that configuration, dL0 provides an additional safeguard, primarily on Qwen. The deployed configuration, \ssa{}-c4 with AR20 and dL0, reads $11.7\%$ of the KV cache.}
  \label{tab:dl-ablation-full}
  \begin{tabular}{lccccc}
    \toprule
    Configuration & Llama & Mistral & Qwen & Avg & $\Delta$ vs.\ dense \\
    \midrule
    Dense & $42.57$ & $42.73$ & $40.18$ & $41.83$ & --- \\
    Fixed-5\% (no verification) & $39.00$ & $37.57$ & $32.49$ & $36.35$ & $-5.48$ \\
    \ssa{}-c4 + AR20, no dL0 & $41.98$ & $41.11$ & $35.53$ & $39.54$ & $-2.29$ \\
    \ssa{}-c4 + AR20 + dL0 & $41.87$ & $41.71$ & $38.54$ & $40.70$ & $-1.13$ \\
    \bottomrule
  \end{tabular}
\end{table}

\paragraph{Extending the dense mask on Qwen.}
Because Qwen carries the largest residual gap, a natural question
(raised in review) is whether densifying more early layers helps.
Extending Qwen's dense mask from layer 0 to layers 0--3 recovers only
$+0.42$ points (from $-1.77$ to $-1.35$ on the probe basis,
paired-bootstrap CI $[-1.85, -0.88]$, still excluding zero) at a
$+9.6$-point increase in effective read ($11.3\%\!\to\!20.9\%$), an unfavorable trade.
Early-layer recovery is detection cost the loop has already repaired:
those rejections are caught and recovered, so quality there is already
preserved.
The residual leak instead originates in late-layer false accepts
(layers $\geq 17$ hold $51\%$ of them; \cref{app:bootstrap}), which
early dense layers cannot reach.
Layer-0-only dense therefore remains the deployed policy.

\paragraph{Approximate versus exact recovery.}
\cref{tab:exact-ar-full} compares the production AR20 recovery policy with Exact Recovery, which computes the complement attention and merges it with the draft using the exact decomposition.
Exact Recovery is the clean algebraic reference path, but it is too expensive to use as the default production policy.
Empirically, AR20 matches Exact Recovery on the three-model average: $40.70$ versus $40.65$.
Per-model differences are small (Llama $-0.24$ pt, Mistral $+0.18$ pt, Qwen $+0.22$ pt), consistent with benchmark noise rather than a systematic quality gap.
Thus, the reported system can use the efficient AR20 path without undermining the theory-to-system story: exact recovery is the reference implied by the proposition, while AR20 is the practical approximation validated here.

\begin{table}[t]
  \centering
  \small
  \caption{Exact Recovery versus AR20. AR20 matches Exact Recovery on average, with per-model differences within $0.24$ pt. \ssa{}-c4 reads $11.7\%$ of the KV cache with AR20.}
  \label{tab:exact-ar-full}
  \begin{tabular}{lccccc}
    \toprule
    Recovery policy & Llama & Mistral & Qwen & Avg & $\Delta$ vs.\ dense \\
    \midrule
    Dense & $42.57$ & $42.73$ & $40.18$ & $41.83$ & --- \\
    \ssa{}-c4 + Exact + dL0 & $42.11$ & $41.53$ & $38.31$ & $40.65$ & $-1.18$ \\
    \ssa{}-c4 + AR20 + dL0 & $41.87$ & $41.71$ & $38.54$ & $40.70$ & $-1.13$ \\
    \bottomrule
  \end{tabular}
\end{table}

\begin{table}[t]
  \centering
  \small
  \setlength{\tabcolsep}{4pt}
  \caption{Additional ungated controls with four-step refresh and dense layer 0
  (LongBench, 16 tasks, 50 samples/task). Logged attention-read fractions for
  the 9\% uniform control are 12.04/12.01/12.48\%, for the 10\% uniform
  control 13.02/13.08/13.40\%, and for the offline control
  12.15/12.09/12.54\% (Llama/Mistral/Qwen). This is a scheduling and
  dense-layer diagnostic, not an exact total-compute match.}
  \label{tab:aligned-controls}
  \begin{tabular}{lcccc}
    \toprule
    Configuration & Llama & Mistral & Qwen & Avg \\
    \midrule
    \ssa{}-c4 (reference) & 41.87 & 41.71 & 38.54 & 40.70 \\
    Uniform 9\%, refresh 4, dL0 & 42.39 & 42.06 & 39.13 & 41.19 \\
    Uniform 10\%, refresh 4, dL0 & 42.39 & 41.92 & 39.17 & 41.16 \\
    Offline budget, refresh 4, dL0 & 42.58 & 41.78 & 38.72 & 41.03 \\
    \bottomrule
  \end{tabular}
\end{table}

\paragraph{Offline per-(layer, head) budget.}
The static control in \cref{tab:matched-compute} gives each (layer, KV head) cell the budget \ssa{} spends there in expectation, $b_{l,h} = 5\% + r_{l,h}\cdot 15\%$, where $r_{l,h}$ is the cell's offline recovery rate, the fraction of decode steps whose coverage falls below $\tau=0.90$ at the $5\%$ base budget (four LongBench tasks, five prompts each, $128$ decode steps).
Budgets are clamped to $[5\%, 20\%]$ and rescaled to a matched effective read ($12.1$--$12.5\%$ across the three models), with the same selector, kernels, samples, and dense layer 0 as \ssa{}-c4. This original static control refreshes every 16 steps, while \ssa{}-c4 checks every four steps; the comparison also changes refresh cadence.
The calibration prompts are drawn from the evaluation distribution, which favors the static policy.
The offline configuration scores $+1.36$ points above the uniform Fixed-12\% control ($+0.31$ / $+0.51$ / $+3.25$ on Llama / Mistral / Qwen), and \ssa{}-c4 scores a further $+2.21$ ($+1.89$ / $+3.12$ / $+1.63$); these are configuration-level differences.
A per-step trace characterizes how coverage failures vary over time.
On a two-sample Llama-3.1-8B HotpotQA trace with a fresh selection at every step ($M{=}1$),
chronic cells (one per sample, layer, and head) that fail on at least $80\%$ of steps are $1.6\%$ of cells, none is chronic in both samples, and they account for only $11.1\%$ of failure events, while intermittent cells failing on $5$--$80\%$ of steps account for $86.4\%$, and the set of failing heads overlaps by a Jaccard index of only $0.33$ between consecutive steps.
The failing heads vary substantially between consecutive steps.

\paragraph{Controls with refresh cadence and dense layer aligned.}
To examine the refresh and dense-layer differences in
\cref{tab:matched-compute}, we also evaluate ungated controls with
four-step refresh and dense layer 0 on the same first 50 examples per
task. \Cref{tab:aligned-controls} retains the original \ssa{}-c4
reference and reports these additional configurations separately.
The uniform 9\% control reaches $41.19$ and the offline control
$41.03$, versus $40.70$ for \ssa{}-c4. Their logged attention-read
fractions are somewhat higher, approximately $12.0$--$12.5\%$ versus
$11.7\%$; page-scoring traffic and wall-clock cost are not matched.
Read fractions are pages-weighted over the last-sample decode trace
retained for each task, rather than over all evaluation examples.
These results show that the original $+3.57$ and $+2.21$ margins
cannot be attributed to the coverage gate alone: refresh cadence and
dense-layer placement materially affect quality.

%% file: appendix/budget_fairness.tex
\section{Budget-Fairness: Baselines at 10\% / 15\% / 20\%}
\label{app:budget-fair}

\begin{figure}[t]
\centering
\includegraphics[width=\textwidth]{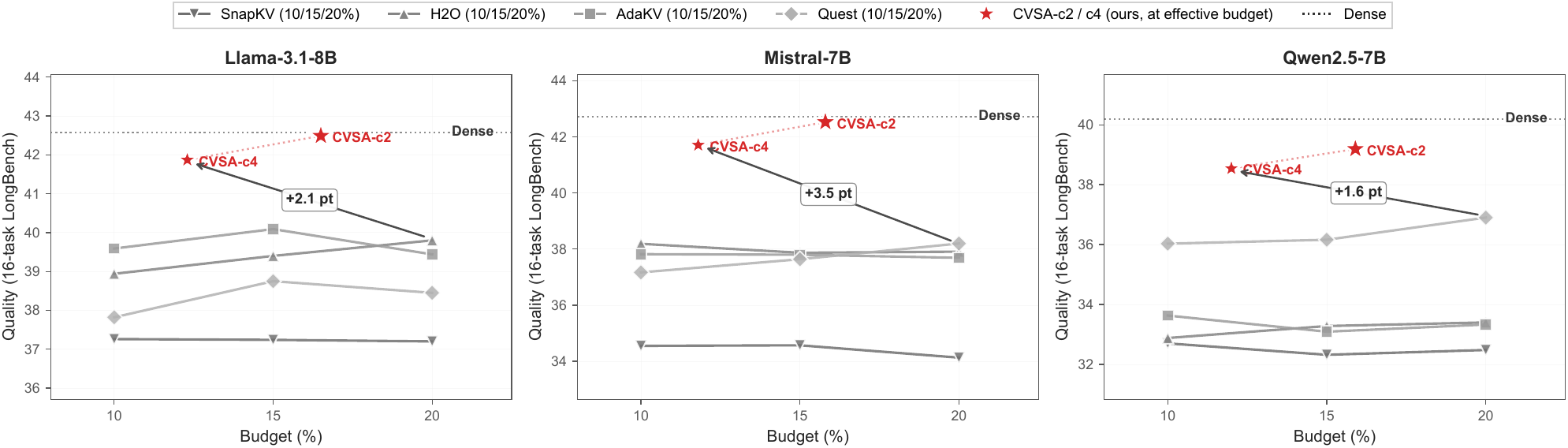}
\caption{Budget-fairness within a single paged-decode implementation: baseline budget swept over $\{10, 15, 20\}\%$ (lines) versus \ssa{}-c2 / \ssa{}-c4 at their measured effective budgets ($\sim$16\% / $\sim$12\%). The baseline curves plateau across the swept range; the diagonal arrows show the remaining configuration-level gaps to \ssa{}-c4. Three 7B models, 16-task LongBench average.}
\label{fig:budget-fairness}
\end{figure}

We sweep each baseline's nominal budget over $\{10, 15, 20\}\%$ within the same paged-decode implementation, on all three 7B models (16-task LongBench).
The defensive question is whether \ssa{}-c4's quality lead disappears once baselines are given comparable or higher budget.
\cref{fig:budget-fairness} shows it does not: every baseline curve is essentially flat across the $10$--$20\%$ range (max single-curve gain is $\leq 1.0$ pt), so doubling the nominal budget does not close the gap to \ssa{}-c4 at $\sim$12\% effective budget.
The diagonal arrow on each panel measures \ssa{}-c4's residual lead over the strongest baseline at $20\%$ budget: $+2.1$ pt on Llama, $+3.5$ pt on Mistral, $+1.6$ pt on Qwen.
Together with \cref{app:isobw} (\ssa{}-c8 at $9.9\%$ effective read exceeds the best paged baseline by $+2.78$ pt), this shows that a larger nominal budget alone does not close the gap for these baselines. Refresh cadence, dense-layer placement, and scoring traffic also differ; these comparisons do not isolate the gate's causal contribution (\cref{app:dl-ar-ablation}).

%% file: appendix/coverage_landscape.tex
\section{Per-Layer Per-Model Coverage Landscape}
\label{app:landscape}

\begin{figure}[t]
\centering
\includegraphics[width=\textwidth]{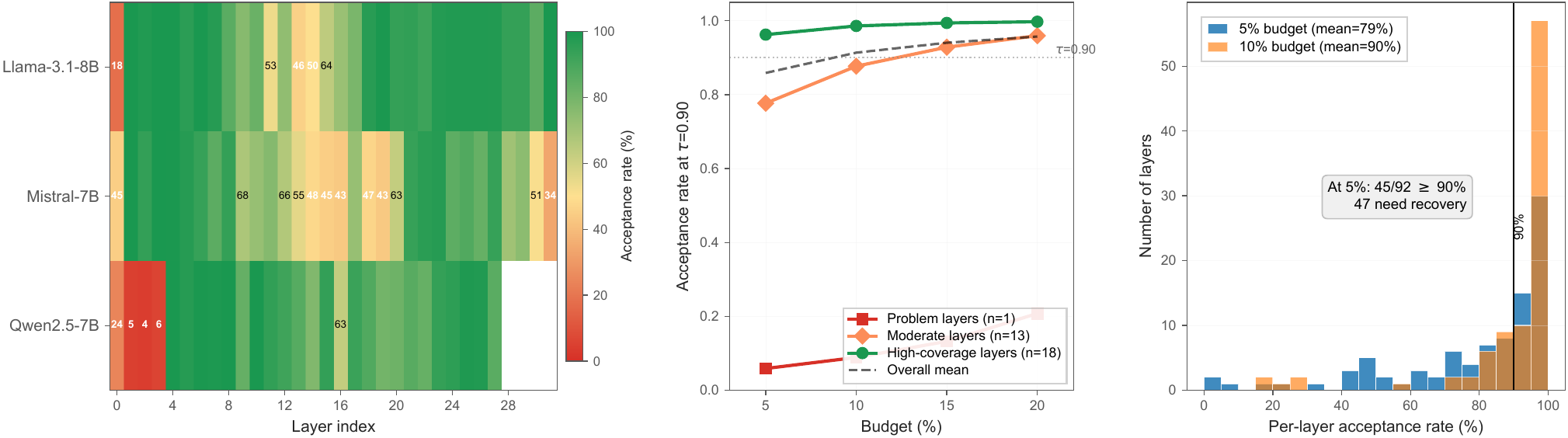}
\caption{Per-layer verifier and coverage landscape on Llama-3.1-8B (32 layers, 4:1 GQA), Mistral-7B-v0.3 (32 layers, 4:1 GQA), and Qwen2.5-7B (28 layers, 7:1 GQA). \emph{(a)} Verifier acceptance rate at $5\%$ budget, $\tau=0.90$: green cells are layer/model pairs that are usually accepted, while red cells identify concentrated reject regions that enter recovery frequently. \emph{(b)} Acceptance rate vs.\ budget on Llama, layers grouped by category: problem layers ($n{=}1$) reach only $21\%$ acceptance even at $4\times$ budget, ruling out budget alone as a fix and motivating the dL0 dense-layer policy; moderate ($n{=}13$) and high-coverage ($n{=}18$) groups saturate by $10$--$15\%$. \emph{(c)} Per-layer acceptance distribution across all $92$ layer-instances: at $5\%$ budget $45/92$ layers accept $\geq 90\%$ of checks; doubling the budget to $10\%$ shifts the distribution rightward but still leaves a low-acceptance tail.}
\label{fig:coverage-heatmap}
\end{figure}

This appendix provides the per-layer verifier diagnostics that underlie the dense-layer policy (\S\ref{sec:dl}) and the coverage-landscape discussion in the ablation studies (\S\ref{sec:ablations}).
\cref{fig:coverage-heatmap} visualises the landscape on three 7B models from three angles: layer-by-model verifier acceptance, budget sweep on Llama, and per-layer acceptance distribution.

\paragraph{Three operational regimes.}
The acceptance heatmap (panel a) and the acceptance histogram (panel c) jointly classify the $92$ layer-instances ($32 + 32 + 28$ across the three models) into three regimes:
\begin{itemize}
    \item \textbf{Problem layers} are low-acceptance regions in panel (a) that enter recovery frequently and, in the lowest-coverage cases, cannot be repaired by a modest budget increase alone.
    \item \textbf{Moderate layers} are neither consistently accepted nor hopeless: they produce a substantial reject rate at the $5\%$ base budget but improve under AR or a larger page budget.
    \item \textbf{High-coverage layers} are accepted almost always, so spending recovery on them would waste bandwidth without improving quality.
\end{itemize}

\paragraph{Budget alone does not fix problem layers.}
Panel (b) sweeps budget over $\{5, 10, 15, 20\}\%$ on Llama and tracks acceptance rate at $\tau=0.90$ separately for each regime.
The problem-layer curve (red squares, $n=1$) reaches only $21\%$ acceptance even at $20\%$ budget (a $4\times$ increase over the default), confirming that no realistic sparse budget will recover these layers, and motivating the dense-layer policy: force these layers to run dense and accept the small fixed cost ($1/32$ of attention compute on a 32-layer model).
Moderate and high-coverage layers in contrast saturate well before $20\%$ budget.

\paragraph{Layer 0 universality.}
Layer 0 is the only layer whose acceptance falls below $50\%$ on all three models (panel a), consistent with the low attention sparsity that Quest~\citep{tang2024quest} reports for the first two layers.
Beyond layer 0, the landscape is model-specific: acceptance also falls below $50\%$ at L1--L3 on Qwen, L13--L14 on Llama, and L14--L16, L18--L19, and L31 on Mistral.
These layers are not forced dense. Every deployed configuration runs only layer 0 dense (\cref{sec:dl}) and leaves the other flagged layers to the gate and approximate recovery; extending Qwen's dense mask to L0--L3
recovers only $+0.42$ points while raising effective read from $11.3\%$ to $20.9\%$ (\cref{app:dl-ar-ablation}).
The diagnostics run automatically on a held-out prompt set, and the dense set is layer 0 on every model, so no per-model tuning is involved.

\paragraph{Implication for verification.}
Of the $92$ layer-instances, $47$ accept fewer than $90\%$ of checks at $5\%$ budget (panel c).
On Llama's moderate layers, a $20\%$ budget raises acceptance from $78\%$ to $96\%$ (panel b); on problem layers it is insufficient, which is why layer 0, the one problem layer shared by all three models, runs dense.
The combined recipe (high-coverage $\to$ accept, moderate $\to$ AR, layer 0 $\to$ dense) is the recipe used throughout the experiments; dense layer 0 raises the three-model average by $1.16$ points, mostly on Qwen (\cref{tab:dl-ablation-full}).

\paragraph{Practitioner rule for keeping a layer dense.}
The recipe above reduces to a two-signal test over quantities the
system already logs offline: keep a layer dense if and only if
\emph{(a)}~its recovery rate at the operating $\tau$ stays near one
(the gate almost always rejects it) \emph{and} \emph{(b)}~its coverage
remains low even after recovery (recovery does not restore the missing
mass).
Condition (a) alone is insufficient, since a layer can reject often
yet be fully repaired by recovery; it is the conjunction that
identifies a layer no sparse budget can serve.
Only layer 0 falls below $50\%$ acceptance on all three 7B models;
Qwen's layers 1--3 reject more often but are repaired by recovery,
and the probe in \cref{app:dl-ar-ablation} prices densifying them
explicitly ($+0.42$ points for $+9.6$ points of read), which the rule correctly declines.

%% file: appendix/recovery_heatmap.tex
\section{Runtime Recovery Concentration}
\label{app:recovery-heatmap}

\cyl{This appendix complements the static coverage landscape in
\cref{app:landscape} with a runtime view of where recovery is actually
spent.  The key point is that recovery is not a hidden dense fallback:
most heads ($77$--$86\%$ per model) are accepted, while the rejected heads concentrate
in a small set of model-specific layers.}

\begin{figure}[t]
\centering
\includegraphics[width=\textwidth]{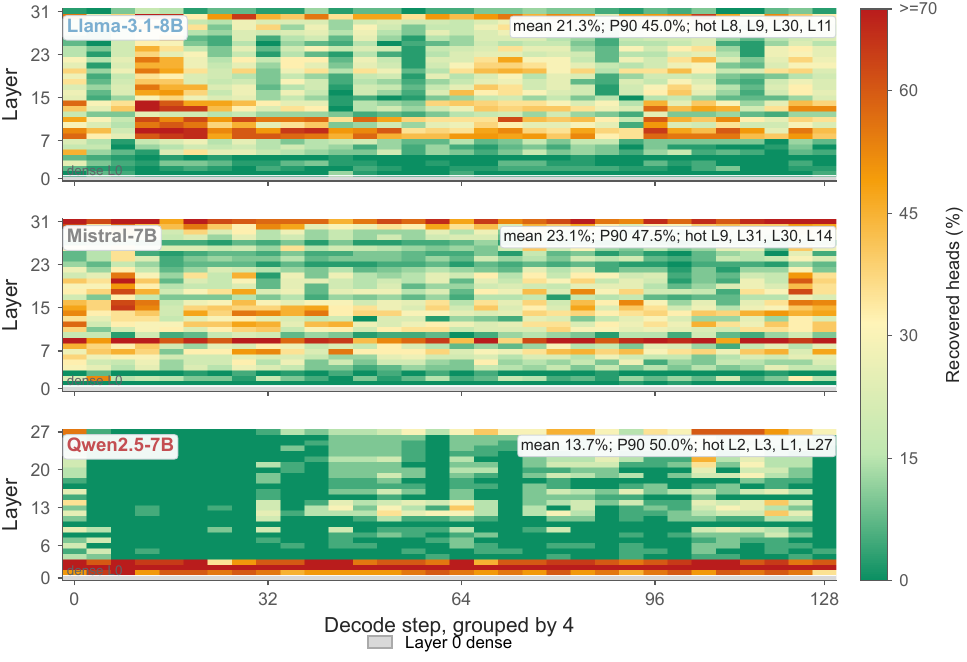}
\caption{\cyl{Where recovery is spent at runtime.  Each panel is a
CVSA-c4 decode trace on HotpotQA, averaged over five samples and
grouped into $M{=}4$ decode-step bins.  Color denotes the fraction of
KV heads in a layer-step bin that fail the coverage check and enter
recovery; layer~0 is gray because the default configuration runs it
dense.  Recovery is concentrated rather than universal: the non-dense
mean recovered-head rates are $21.3\%$ on Llama-3.1-8B, $23.1\%$ on
Mistral-7B, and $13.7\%$ on Qwen2.5-7B, with 90th-percentile
layer-step rates of $45.0\%$, $47.5\%$, and $50.0\%$ respectively.
The hotspots are model-specific (Llama L8/L9/L30/L11, Mistral
L9/L31/L30/L14, Qwen L2/L3/L1/L27), supporting the claim that
verification routes recovery to a small low-coverage tail while most
heads ($77$--$86\%$ per model) remain on the accepted sparse path.}}
\label{fig:recovery-heatmap}
\end{figure}

\cyl{%
\Cref{fig:recovery-heatmap} should be read as a runtime complement to
the static coverage landscape in \cref{app:landscape}: the landscape
identifies which layers have low coverage at a given budget, while the
heatmap shows where recovery is actually spent during generation.
Color saturation encodes the fraction of KV heads in each layer-step
bin that fail the coverage check and enter recovery.
The key finding is that recovery is selective rather than global.
Llama and Mistral show late-layer and mid-layer hotspots (L8/L9/L30
on Llama, L9/L31/L30 on Mistral), while Qwen concentrates rejection
in the first few non-dense layers (L2/L3), consistent with the
low early-layer acceptance on Qwen in \cref{app:landscape}.
The non-dense mean recovered-head fractions are 21.3\% (Llama),
23.1\% (Mistral), and 13.7\% (Qwen), with 90th-percentile
layer-step rates of 45--50\%.
Because these hotspots occupy a small portion of the layer-step
plane, the amortized bandwidth cost of recovery stays close to the
base sparse budget, which explains why the throughput reported in
the main paper is not erased by the verify-then-correct loop.
This pattern supports the central design assumption in
\cref{sec:coverage}: sparse-attention failures are concentrated
rather than uniform, making selective recovery more efficient than
uniform budget inflation.%
}


%% file: appendix/ruler.tex
\section[Full RULER Numbers]{Per-Task $\times$ Per-Model $\times$ Per-Length Numbers (RULER)}
\label{app:rulerfull}
The full RULER grid behind \cref{fig:ruler-main} covers all six tasks
at 4K--32K for every method, plus 64K and 128K runs for dense, \ssa{},
and Fixed-10\%, and for the three eviction baselines on the
\cref{fig:ruler-main} tasks (MV-NIAH, MQ-NIAH, CWE).
\Cref{tab:ruler-summary} condenses the grid into the mean gap to dense,
and \cref{tab:ruler-full-llama,tab:ruler-full-mistral,tab:ruler-full-qwen}
give every cell.
On the $72$ cells every method covers (six tasks at 4K--32K on three
models), \ssa{}-c4 is $0.83$ points below dense and the strongest
baseline, H2O, is $8.70$ points below.
The tables are generated by \texttt{eval/build\_ruler\_table.py} from
the JSON files under \texttt{results/}.

%% file: appendix/llama70b.tex
\section{Llama-3.1-70B Detailed Results}
\label{app:70b}

This appendix expands \cref{tab:70b} with the full per-task breakdown on Llama-3.1-70B-Instruct (4$\times$H200) and explains the methodology behind the 7-task subset.

\paragraph{Why 7 tasks and not 8.}
Our 8-task English subset of LongBench includes \texttt{samsum}, which is verbose dialogue summarization with multiple acceptable end-of-turn markers.
Our custom multi-GPU paged-decode generation loop does not reproduce HuggingFace \texttt{generate}'s stopping behavior on this open-ended summarization task, even after matching the multiple EOS variants used by Llama-3.1-Instruct.
On 70B at long context this divergence inflates samsum's verbose tail and produces score artifacts (paged: $\sim$19 ROUGE-L; HF: $\sim$45) that are not method-attributable.
We exclude samsum on 70B for this reason and report 7-task averages throughout this appendix; the same task is included for the 7B models, where the divergence does not manifest.

\paragraph{Per-task breakdown.}
\cref{tab:70b-pertask} reports per-task scores for paged dense, \ssa{}-c2, \ssa{}-c4, KVPress AdaKV, and KVPress SnapKV ($n=30$ samples per task, greedy decoding, 10\% nominal budget for baselines, $5\%$ budget plus AR$=20\%$ and dL0 for \ssa{}).
\ssa{}-c2 matches or exceeds paged dense on 4 of 7 tasks (hotpotqa, triviaqa, multifieldqa\_en, trec) and is within 1 pt on the remaining three; the bootstrap 95\% CI on the 7-task average is $[-0.79, +1.95]$, overlapping zero.
\ssa{}-c4 degrades on extractive tasks (lcc $-3.16$, hotpotqa $-2.26$, triviaqa $-3.67$) because error accumulates over 80 layers when the check interval is too long relative to model depth, which motivates the $M{=}2$ guideline for 70B-scale models.

\begin{table}[t]
  \centering
  \small
  \caption{Llama-3.1-70B-Instruct per-task scores on the 7-task LongBench subset (samsum excluded; see text). KVPress baselines run in HuggingFace SDPA, while \ssa{} runs in our paged-decode loop; a 2.48-pt cross-pipeline drift on the dense reference (paged $52.02$ vs.\ KVPress $54.50$) sets a noise floor for cross-implementation comparisons; KVPress baselines therefore appear $2.48$ points stronger than they would under the paged reference. Within each pipeline, \ssa{}-c2 statistically matches paged dense (CI $[-0.79, +1.95]$); \ssa{}-c4 underperforms because $M{=}4$ is too long for 80 layers.}
  \label{tab:70b-pertask}
  \begin{tabular}{lccccc}
    \toprule
    Task                  & Paged dense & \ssa{}-c2     & \ssa{}-c4 & KVPress AdaKV & KVPress SnapKV \\
    \midrule
    hotpotqa              & $43.10$     & $\mathbf{45.14}$ & $40.84$ & $48.37$       & $48.37$        \\
    triviaqa              & $90.37$     & $\mathbf{91.40}$ & $86.70$ & $92.00$       & $90.00$        \\
    qasper                & $58.85$     & $58.04$       & $55.31$ & $56.90$       & $55.46$        \\
    multifieldqa\_en      & $55.79$     & $\mathbf{58.20}$ & $54.22$ & $60.79$       & $58.78$        \\
    trec                  & $66.67$     & $\mathbf{66.67}$ & $66.67$ & $66.67$       & $66.67$        \\
    lcc                   & $23.93$     & $23.67$       & $20.77$ & $27.90$       & $28.50$        \\
    narrativeqa           & $25.46$     & $25.27$       & $23.80$ & $25.96$       & $25.55$        \\
    \midrule
    7-task average        & $52.02$     & $\mathbf{52.63}$ & $49.76$ & $54.08$       & $53.33$        \\
    $\Delta$ vs.\ paged dense & ---     & $\mathbf{+0.60}$ & $-2.27$ & $+2.06$       & $+1.31$        \\
    \bottomrule
  \end{tabular}
\end{table}

\paragraph{Cross-pipeline reading.}
The $+2.06$ pt apparent gain of KVPress AdaKV over paged dense in \cref{tab:70b-pertask} reflects the cross-pipeline drift, not method gain: the KVPress dense reference itself is $54.50$, $+2.48$ pt above paged dense, larger than the AdaKV margin.
The within-pipeline reading is: KVPress AdaKV is $\Delta = -0.41$ vs.\ KVPress dense, while \ssa{}-c2 is $\Delta = +0.60$ vs.\ paged dense; each method is evaluated against its own pipeline.

\paragraph{Recommendation for 70B-scale models.}
On 80-layer models, error accumulates faster than at 32-layer scale; we recommend $M{=}2$ (\ssa{}-c2) for $\geq$70B and keep $M{=}4$ as the 7B default.
The $M$-vs.-depth trade-off is not yet captured by a closed-form policy and is a candidate for follow-up work.

%% file: appendix/roofline.tex
\section{Bandwidth-Roofline Derivation}
\label{app:roofline}

The decode-step throughput speedups reported in \S\ref{sec:latency} are predicted by a simple bandwidth-roofline model.
This appendix derives the model, instantiates the parameters per architecture, and reports measured-vs-predicted speedups across $\{$model, batch$\}$ pairs.

\paragraph{Setup.}
At a single decode step, the GPU must read (i) all model weights $W$ once and (ii) the KV cache for every sequence in the batch.
Let
\begin{align*}
W_{\text{bytes}}     &= \text{total model-weight bytes (bf16)},  \\
\mathrm{KV}_{\text{bytes}} &= \text{KV-cache bytes for one sequence at the current context length},  \\
B              &= \text{batch size},\\
\eta             &= \text{effective KV-read fraction relative to dense.}
\end{align*}
For dense attention $\eta=1$; for \ssa{}-c4, $\eta \approx 0.12$.

The dense decode-step is then bandwidth-bound by $W_{\text{bytes}} + B \cdot \mathrm{KV}_{\text{bytes}}$ HBM bytes per step.
\ssa{}-c4 reads the same weights but only $\eta$ of the KV cache, so its total HBM traffic per step is $W_{\text{bytes}} + B \cdot \eta \cdot \mathrm{KV}_{\text{bytes}}$.

\paragraph{Speedup formula.}
Define $f_{\text{kv}}$ as the KV-cache fraction of total decode-step HBM traffic for dense:
\begin{equation}
f_{\text{kv}} = \frac{B \cdot \mathrm{KV}_{\text{bytes}}}{W_{\text{bytes}} + B \cdot \mathrm{KV}_{\text{bytes}}}.
\label{eq:fkv}
\end{equation}
Because dense decode is bandwidth-bound, the speedup of \ssa{}-c4 over dense is the inverse ratio of HBM traffic:
\begin{align}
\text{speedup}
&= \frac{W_{\text{bytes}} + B \cdot \mathrm{KV}_{\text{bytes}}}{W_{\text{bytes}} + B \cdot \eta \cdot \mathrm{KV}_{\text{bytes}}} \nonumber \\
&= \frac{1}{1 - f_{\text{kv}} + f_{\text{kv}} \cdot \eta},
\label{eq:speedup}
\end{align}
where the second line follows by dividing both numerator and denominator by $W_{\text{bytes}} + B \cdot \mathrm{KV}_{\text{bytes}}$ and substituting Eq.~\eqref{eq:fkv}.

\paragraph{Parameter values.}
\cref{tab:roofline-spec} lists the architecture parameters used to compute $W_{\text{bytes}}$ (per-layer Q + K + V + O projections plus the three-matrix MLP, summed over layers, in bf16) and $\mathrm{KV}_{\text{bytes}}$ (16-token pages, $\mathrm{kv\_heads} \cdot \mathrm{head\_dim}$ bytes per page-side-K-or-V, summed over layers and the running 8K context).
Qwen2.5-7B's 7:1 GQA ratio gives the smallest KV cache per layer relative to its larger MLP (intermediate dim $18944$ vs.\ $14336$ on Llama / Mistral), so its $f_{\text{kv}}$ is the lowest of the three models at every batch size and the bandwidth-only model predicts it the smallest speedup; its measured speedup is nonetheless the largest, for the reasons given under \cref{tab:roofline-pred-meas}.

\begin{table}[t]
  \centering
  \small
  \caption{Architecture specs used in the roofline model. KV-cache bytes scale linearly with $\mathrm{kv\_heads}$ and context length; MLP bytes (which dominate weight reads at small $B$) scale with $\mathrm{hidden} \times \mathrm{intermediate}$.}
  \label{tab:roofline-spec}
  \begin{tabular}{lcccccc}
    \toprule
    Model        & Layers & Q/KV heads & Head dim & Hidden & MLP inter. \\
    \midrule
    Llama-3.1-8B   & 32 & 32 / 8 & 128 & 4096 & 14336 \\
    Mistral-7B-v0.3 & 32 & 32 / 8 & 128 & 4096 & 14336 \\
    Qwen2.5-7B     & 28 & 28 / 4 & 128 & 3584 & 18944 \\
    \bottomrule
  \end{tabular}
\end{table}

\begin{figure}[t]
\centering
\includegraphics[width=0.85\textwidth]{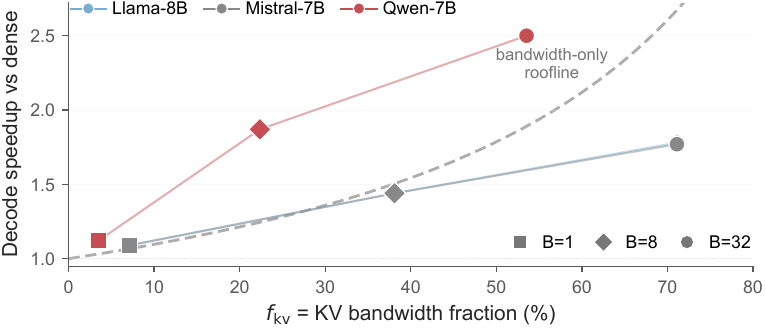}
\caption{Decode-step speedup vs.\ KV bandwidth fraction $f_{\text{kv}}$. The dashed grey curve is the bandwidth-only prediction $1/(1 - f_{\text{kv}} + f_{\text{kv}} \cdot \eta)$ with $\eta = 0.12$; markers are measured \ssa{}-c4 speedups across three 7B models and three batch sizes ($B \in \{1, 8, 32\}$, 8K context). The model captures the qualitative trend (speedup grows with $f_{\text{kv}}$, i.e.\ with both context length and batch size); Qwen's elevated $B \geq 8$ measurements reflect dL0 overhead amortization that is not in the simple roofline.}
\label{fig:roofline}
\end{figure}

\paragraph{Predicted vs.\ measured speedup.}
\cref{tab:roofline-pred-meas} reports the per-(model, batch) breakdown at 8K context with $\eta = 0.12$ (\ssa{}-c4's measured average effective budget).
\cref{fig:roofline} plots all nine points against the theoretical speedup curve (Eq.~\eqref{eq:speedup}).

\begin{table}[t]
  \centering
  \small
  \caption{Predicted speedup from Eq.~\eqref{eq:speedup} (using $\eta = 0.12$, \ssa{}-c4's $11.7\%$ effective read on LongBench, rounded) versus measured speedup at 8K context, three batch sizes. Qwen consistently exceeds the bandwidth-only prediction at $B \geq 8$ because dL0 (force layer 0 to dense) reduces effective per-decode latency disproportionately on its 7:1 GQA architecture, and the Triton-kernel index-computation overhead is also smaller for the smaller KV per layer; Llama and Mistral at $B{=}32$ fall short of the bandwidth ceiling because non-attention overheads (LSE merge, page scoring) become a larger fraction of the shrunk attention time. The model is intended as a directional predictor, not an exact fit.}
  \label{tab:roofline-pred-meas}
  \begin{tabular}{lrccc}
    \toprule
    Model           & $B$ & $f_{\text{kv}}$ (\%) & Predicted (Eq.~\ref{eq:speedup}) & Measured \\
    \midrule
    Llama-3.1-8B   &  1 &  $7.1$  & $1.07\times$ & $1.09\times$ \\
                    &  8 & $38.1$  & $1.50\times$ & $1.44\times$ \\
                    & 32 & $71.1$  & $2.67\times$ & $1.78\times$ \\
    \midrule
    Mistral-7B-v0.3 &  1 &  $7.1$  & $1.07\times$ & $1.09\times$ \\
                    &  8 & $38.1$  & $1.50\times$ & $1.44\times$ \\
                    & 32 & $71.1$  & $2.67\times$ & $1.77\times$ \\
    \midrule
    Qwen2.5-7B     &  1 &  $3.5$  & $1.03\times$ & $1.12\times$ \\
                    &  8 & $22.4$  & $1.24\times$ & $1.87\times$ \\
                    & 32 & $53.5$  & $1.89\times$ & $\mathbf{2.50\times}$ \\
    \bottomrule
  \end{tabular}
\end{table}

\paragraph{Implications for paper claims.}
The roofline model justifies the qualitative scaling claims in \S\ref{sec:latency}: as context grows, $\mathrm{KV}_{\text{bytes}}$ grows linearly in context while $W_{\text{bytes}}$ stays fixed, so $f_{\text{kv}} \to 1$ and speedup approaches the asymptotic ceiling $1 / \eta \approx 8.3\times$.
At short context and small batch, weight reads dominate ($f_{\text{kv}} \le 7\%$ at 8K, $B{=}1$), so the gain is small regardless of $\eta$ (measured $1.09$--$1.12\times$).

%% file: appendix/throughput.tex
\section{Long-Context Throughput Tables}
\label{app:longctx-table}

This appendix expands the headline numbers in \cref{tab:long-ctx} (\S\ref{sec:latency}) with measurements across three 7B models at long context.
All values are median decode-step speedup vs.\ dense ($\mathrm{dense\_ms} / \mathrm{c4\_ms}$, median over timed decode steps) on the paged-decode pipeline.
This statistic differs from aggregate tokens/s when scoring steps have
higher latency: at Qwen 128K, $B=8$, the median-step ratio is $6.21\times$,
while the logged tokens/s ratio is $115.60/24.86 = 4.65\times$.
Llama and Mistral are measured on an H100 (80 GB, \S\ref{sec:setup}); at $B{=}8$ and $\geq$65K their bf16 KV cache plus weights exceeds this memory ($\approx$85 GB at 65K), so these cells are not reported.

\begin{table}[t]
  \centering
  \small
  \caption{\ssa{}-c4 median decode-step speedup vs.\ dense across three 7B models, three context lengths, two batch sizes; Llama and Mistral run on H100, Qwen on H200. ``---'': exceeds the H100's 80 GB (Llama / Mistral, $B{=}8$, $\geq 65$K). All speedups are medians over the timed decode steps on the paged-decode pipeline. \ssa{}-c4 reads $11.7\%$ of the KV cache on LongBench.}
  \label{tab:longctx-throughput}
  \begin{tabular}{lcccccc}
    \toprule
    & \multicolumn{3}{c}{B=1 (single-stream)} & \multicolumn{3}{c}{B=8 (batched)} \\
    \cmidrule(lr){2-4} \cmidrule(lr){5-7}
    Model           & 32K   & 65K   & 128K  & 32K   & 65K   & 128K  \\
    \midrule
    Llama-3.1-8B   & $1.12\times$ & $1.15\times$ & $1.19\times$ & $1.55\times$ & ---           & ---           \\
    Mistral-7B-v0.3 & $1.13\times$ & $1.16\times$ & $1.19\times$ & $1.56\times$ & ---           & ---           \\
    Qwen2.5-7B     & $1.40\times$ & $1.86\times$ & $2.68\times$ & $3.07\times$ & $4.44\times$ & $\mathbf{6.21\times}$ \\
    \bottomrule
  \end{tabular}
\end{table}

\paragraph{Speedup grows with context.}
Within every (model, batch) pair, speedup is monotonic in context length, consistent with the bandwidth-roofline prediction (\cref{app:roofline}): $f_{\text{kv}}$ rises with context, so attention dominates more of the decode-step cost and the $\eta = 0.12$ KV-read reduction translates into a larger end-to-end win.

\paragraph{Speedup grows with batch.}
At every (model, context) pair, B=8 is faster than B=1 because batching amortizes the fixed weight reads across more queries, shifting $f_{\text{kv}}$ upward (Eq.~\ref{eq:fkv}).
For example on Qwen at 128K: B=1 gives $2.68\times$, B=8 gives $6.21\times$.

\paragraph{Qwen has the largest speedup, though not the largest KV share.}
Qwen's 7:1 GQA combined with its larger MLP intermediate ($18944$ vs.\ $14336$ on Llama / Mistral) produces the smallest KV-cache-to-weights ratio and hence the lowest $f_{\text{kv}}$ of the three models (\cref{tab:roofline-pred-meas}). Its larger speedup comes from what the bandwidth-only roofline omits: the dense layer 0 costs it disproportionately little per decode step, and the Triton kernels' index-computation overhead is smaller for its smaller per-layer KV.
Llama and Mistral cluster within $0.02\times$ of each other at every measured cell because their architectural specs are nearly identical (\cref{tab:roofline-spec}).


\paragraph{Per-kernel latency breakdown.}
To locate where decode-step time is spent, and in particular to
measure the cost of the verifier itself, we profile the four
attention-path kernels with per-kernel CUDA timing on Llama-3.1-8B at
batch $8$, 65K context on one H200, over $67$ decode steps (\cref{tab:kernel-breakdown}).
The coverage check, i.e.\ the verifier, accounts for under $2\%$ of
attention-path latency: it reuses the page scores that page scoring
already produces and launches no additional attention pass.
The material costs are page scoring (every $M{=}4$ steps) and
conditional recovery.
Mistral-7B reproduces all four figures within $8\%$.

\begin{table}[t]
  \centering
  \small
  \caption{Per-kernel decode-step latency on Llama-3.1-8B (batch $8$, 65K context, one H200, $67$ decode steps). The verifier (coverage check) is under $2\%$ of attention-path time because it consumes quantities the kernels already compute; page scoring and conditional recovery occur within the timed decode steps, although their cost is not fully represented by a median-step speedup. Mistral-7B reproduces each figure within $8\%$. \ssa{}-c4 reads $11.7\%$ of the KV cache on LongBench.}
  \label{tab:kernel-breakdown}
  \begin{tabular}{lcc}
    \toprule
    Attention-path kernel & ms / decode step & Share \\
    \midrule
    Coverage check (verifier)   & $0.45$ & $<2\%$ \\
    Page scoring                & $8.81$ & ${\sim}39\%$ \\
    Conditional recovery        & $9.61$ & ${\sim}42\%$ \\
    Sparse draft attention      & $3.80$ & ${\sim}17\%$ \\
    \bottomrule
  \end{tabular}
\end{table}

\paragraph{CUDA Graph compatibility.}
The recovery decision is computed on the device: the coverage kernel
emits a per-head accept mask from quantities already resident in GPU
memory, and the merge of draft and recovery outputs is a per-head
masked select, so the mechanism requires no host-side round trip.
The current implementation takes one host-side branch per scoring
step, an early exit that skips the recovery kernel when every head
accepts, which is an implementation convenience rather than a
requirement of the method.
A graph-capturable variant launches the recovery kernel
unconditionally with the accept mask applied as predication, at small
added cost because recovery already runs on nearly every scoring step
at the deployed operating point, or places the branch in a conditional
graph node (available since CUDA~12.4).
Reuse steps, three of every four at $M{=}4$, decode a cached
fixed-shape selection and are directly capturable.
All speedups in this paper were measured in a single eager pipeline
with no graph capture for any method, a setting that if anything
understates \ssa{}'s relative speed because \ssa{} issues the most
kernel launches per decode step.

%% file: appendix/iso_bandwidth.tex
\section{\ssa{}-c8 at a Matched Effective Budget}
\label{app:isobw}

This appendix elaborates on the matched-budget claim in \S\ref{sec:quality}: \ssa{}-c8's $\Delta=-2.04$ at $9.9\%$ effective budget beats the strongest paged baseline by $+2.78$ pt at a matched effective budget, showing a quality advantage at a comparable attention-read budget for these configurations.

\paragraph{Effective-budget definition (recap).}
Effective budget is the fraction of KV pages read by attention relative to dense, including AR expansion and dense layer 0; page-scoring reads are counted separately (\cref{app:selector-traffic}).
For \ssa{}-c8: $5\%$ base budget, $\tau=0.90$, $M=8$, AR$=20\%$ on rejected heads.
The accept rate at $\tau=0.90$ is $\sim$50\%, so weighted across head-steps the effective budget lands at $9.9\%$, marginally below the baselines' $10\%$ nominal budget, allowing a matched-budget comparison.

\paragraph{Quality at iso-bandwidth.}
\cref{tab:isobw} compares \ssa{}-c8 (effective budget $9.9\%$) against the four baselines (nominal budget $10\%$) on three 7B models, LongBench 16-task average.
\ssa{}-c8 retains the verification advantage: at a lower attention-read fraction, it still beats every paged baseline on the three-model average and matches or beats every baseline on every individual model.

\begin{table}[t]
  \centering
  \small
  \caption{\ssa{}-c8 at a matched effective budget: $9.9\%$ effective budget vs.\ baselines at $10\%$ nominal budget (SnapKV and AdaKV via KVPress; H2O and Quest paged). The $+2.78$ pt margin over the strongest paged baseline (Quest) holds at a marginally lower attention-read fraction. Comparing against canonical KVPress AdaKV and SnapKV at $10\%$ (\cref{tab:longbench}), \ssa{}-c8's $\Delta=-2.04$ is also better than KVPress AdaKV's $\Delta=-3.05$.}
  \label{tab:isobw}
  \begin{tabular}{lcccc}
    \toprule
    Method                & Eff.\ budget & 3-model avg  & $\Delta$ vs.\ dense   & $\Delta$ vs.\ Quest \\
    \midrule
    Dense                 & $100\%$      & $41.83$      & ---                   & ---           \\
    \midrule
    \textbf{\ssa{}-c8 (ours)} & $\mathbf{9.9\%}$  & $\mathbf{39.79}$ & $\mathbf{-2.04}$ & $\mathbf{+2.78}$ \\
    \midrule
    SnapKV (KVPress)      & $10\%$       & $38.56$      & $-3.27$               & $+1.55$       \\
    AdaKV (KVPress)       & $10\%$       & $38.78$      & $-3.05$               & $+1.77$       \\
    H2O (paged)           & $10\%$       & $36.67$      & $-5.16$               & $-0.34$       \\
    Quest (paged)         & $10\%$       & $37.01$      & $-4.82$               & ---           \\
    \bottomrule
  \end{tabular}
\end{table}

\paragraph{Implication.}
The attention-read comparison and the budget-fairness sweep
(\cref{app:budget-fair}) show that increasing the evaluated baselines'
nominal budgets does not close their quality gap to \ssa{}.
Attention-page counts do not match total compute or refresh cadence;
\cref{app:dl-ar-ablation} examines the latter together with dense-layer placement.

%% file: appendix/selector_traffic.tex
\section{Selector Compatibility and Traffic Accounting}
\label{app:selector-traffic}

This appendix measures compatibility with alternative page selectors
and accounts for the traffic each configuration consumes.
Unless noted otherwise, all numbers use the same LongBench 16-task protocol, samples, kernels,
and scorer as the main results; effective KV-read is pages-weighted
over the last-sample decode trace retained for each task in the execution logs.

\paragraph{Per-configuration traffic accounting.}
\cref{tab:traffic-accounting} reports effective read and quality for
three configurations that share our pipeline: a pre-hoc top-$p$
selector alone, the same selector with the unchanged coverage gate,
and \ssa{}-c4.
The pre-hoc selector reads more than \ssa{}-c4 ($12.73\%$ versus
$11.69\%$ on the three-model average) yet scores $9.0$ points lower
($31.69$ vs.\ $40.70$); composing the gate on that selector recovers
$+3.67$ (to $35.36$) while still reading more than \ssa{}-c4.
Quality does not follow from traffic alone.

\begin{table}[t]
  \centering
  \small
  \caption{Per-configuration effective KV-read (\%) and LongBench quality across three 7B models. \ssa{}-c4 sets a coverage level rather than a per-head budget, so its traffic is an outcome ($\sim$11.7\% on every model with no per-model tuning), not a tuned input. The pre-hoc selector consumes more traffic at lower quality; adding the gate improves it by $+3.67$ but does not reach \ssa{}-c4.}
  \label{tab:traffic-accounting}
  \begin{tabular}{lcccc}
    \toprule
    Configuration (read\% / quality) & Llama & Mistral & Qwen & 3-model \\
    \midrule
    Pre-hoc top-$p$ (alone)          & 14.42 / 35.12 & 13.55 / 32.79 & 10.21 / 27.16 & 12.73 / 31.69 \\
    Pre-hoc top-$p$ + coverage gate  & 16.88 / 38.94 & 16.94 / 36.21 & 13.64 / 30.93 & 15.82 / 35.36 \\
    \ssa{}-c4 (verified top-$k$)     & 11.67 / 41.87 & 11.74 / 41.71 & 11.66 / 38.54 & 11.69 / 40.70 \\
    \bottomrule
  \end{tabular}
\end{table}

\paragraph{Selector-agnostic verification.}
\cref{tab:selector-agnostic} evaluates the coverage gate under three
selector families beyond the deployed exact top-$k$, with the gate and
recovery unchanged.
The gate detects whatever each selector misses and routes recovery
there, so a weaker selector degrades gracefully rather than failing
silently, and exchanging the allocation rule leaves the outcome
essentially unchanged in the gated configurations. Ungated comparisons
also differ in scheduling; they are configuration-level comparisons.

\begin{table}[t]
  \centering
  \small
  \caption{Coverage gate under alternative page selectors (LongBench 16-task average, three 7B models). The gated configurations improve on the corresponding ungated configurations on every model ($+2.19$ / $+1.74$ / $+1.37$ and $+3.82$ / $+3.42$ / $+3.77$). The AdaKV-style allocator is measured against \ssa{}-c4's $40.70$ at the same effective read ($+0.10$ / $-0.31$ / $+0.01$ per model), so exchanging the allocation rule moves quality by $0.06$ points. \ssa{}-c4 reads $11.7\%$ of the KV cache.}
  \label{tab:selector-agnostic}
  \begin{tabular}{lccc}
    \toprule
    Selector (gate and recovery unchanged) & Without gate & With gate & Change \\
    \midrule
    Quest-style min/max indexer       & $37.01$ & $38.77$ & $+1.76$ \\
    Pre-hoc top-$p$ (adaptive budget) & $31.69$ & $35.36$ & $+3.67$ \\
    AdaKV-style cross-head allocation & ---     & $40.64$ & $-0.06$ \\
    \bottomrule
  \end{tabular}
\end{table}
The top-$p$ rows of \cref{tab:traffic-accounting} expand the selection
until an estimated mass threshold is met at page granularity, using an
fp16 estimator stronger than the 4-bit estimator of the released
Twilight implementation, so that family is evaluated in the most
favorable setting we could construct for it.

\paragraph{Reproducing Tactic.}
Tactic~\citep{zhu2025tactic} has no public reference implementation, so
we reimplemented it from the paper: $k$-means over keys at prefill,
decode-time cluster ranking by centroid inner product, exact scores for
the leading clusters, a $y = a/x + b$ fit for the remaining tail, and
accumulation until the estimated attention mass reaches the target,
unioned across each GQA group.
We validated it against the two operating points the paper reports for
Llama-3.1-8B, on the six tasks it evaluates, against a criterion fixed
before the runs: the reproduction reaches $-0.54$ against a published
$-1.37$ at a target mass of $0.70$, and $-0.59$ against a published
$-0.34$ at $0.90$.

One departure from the text is necessary, and we state it plainly: we
force the first four tokens into every selection.
The paper does not mention attention sinks, and without them the
reproduction misses the published low-target point by $3.35$ while
matching the high-target one, since at a low target too few clusters are
selected for the sink to be included incidentally.
Adding the sinks raises quality by $4.19$ and \emph{lowers} read by
$1.81$ points, and brings both operating points inside the criterion, so
the departure strengthens and cheapens the method rather than weakening
it.
A recent-window variant was also tried and is not used: it did not
improve on sinks alone, and inventing further machinery for a competing
method is its own kind of infidelity.

\begin{table}[t]
  \centering
  \small
  \caption{The validated Tactic reproduction under this paper's protocol (Llama-3.1-8B, LongBench 16-task average, same samples, kernels and scorer as \cref{tab:longbench}). KV read is the share of the cache each method names; pages touched is the share a paged serving stack must fetch to deliver it. The two coincide for \ssa{}, which reads pages directly. Both columns count attention traffic only; scoring traffic is larger for \ssa{} than for Tactic and is accounted for in the text.}
  \label{tab:tactic-comparison}
  \begin{tabular}{lcccc}
    \toprule
    Method & Quality & $\Delta$ & KV read & Pages touched \\
    \midrule
    Dense                       & $42.57$ & ---     & $100\%$   & $100\%$ \\
    \textbf{\ssa{}-c4}          & \cvstabbest{$41.87$} & \cvstabbest{$-0.71$} & \cvstabbest{$11.67\%$} & \cvstabbest{$11.67\%$} \\
    Tactic (reproduced)         & $41.82$ & $-0.76$ & $22.98\%$ & $48.60\%$ \\
    \quad + coverage gate       & $42.49$ & $-0.08$ & $31.73\%$ & $59.21\%$ \\
    \bottomrule
  \end{tabular}
\end{table}

\cref{tab:tactic-comparison} places the validated reproduction under this
paper's protocol.
Quality is matched: $0.05$ points separate the two methods, well inside
run-to-run variation, so neither is the more accurate.
They differ in what they read to get there.
Tactic selects individual tokens, and those tokens are scattered by
construction, because its clusters group keys by similarity rather than
by position.
Charged at page granularity, the unit a paged serving stack fetches and
the setting this paper targets, the $22.98\%$ of tokens it names occupy
$48.60\%$ of pages; \ssa{} reads pages directly, so its two figures
coincide at $11.67\%$.
Even before page granularity is applied, the reproduction names roughly
twice the cache share for the same quality.

These columns count attention traffic only, and the scoring traffic that
each method needs to produce its selection runs the other way.
\ssa{} scores pages from a full key scan amortized over the $M{=}4$
reuse window, about $12.5\%$ of the cache per step, while Tactic's
centroid ranking, leading-cluster scores and tail fit come to roughly
$5\%$.
Including both, the totals are $24.2\%$ for \ssa{} against $28.0\%$ for
the reproduction where token gather is free, and $53.6\%$ on a paged
stack.
The advantage in total traffic is therefore a factor of $1.2$ in the
first setting and $2.2$ in the second, not the ratio the attention
columns alone would suggest.

Composing the unchanged coverage gate on top of the reproduction is the
result that bears on this paper's claim.
It moves Tactic from $-0.76$ to $-0.08$, closing most of the remaining
distance to dense, and the verifier rejects $64.85\%$ of the checks it
performs: over sixteen tasks the pre-hoc mass target is missed about two
times in three, and repairing those steps is worth $0.68$ points at the
cost of the traffic the repair consumes.
Verification is therefore complementary to this selector rather than an
alternative to it, which is the same relationship the audit below finds
for the deployed member of the family.
The comparison is at this paper's protocol; the six validation tasks are
Tactic's own selection and are weighted toward retrieval and extractive
question answering, where token-granular selection is at its strongest.

\paragraph{Auditing a pre-hoc adaptive budget at its source.}
The top-$p$ rows above instantiate the pre-hoc principle inside our own
pipeline; it can also be audited where it is actually deployed.
We ran the released Twilight implementation~\citep{lin2025twilight}
unchanged, at its own default configuration (Quest selector, token
budget $8192$, hierarchical top-$p$ pruning at $p{=}0.85$, 4-bit
min/max weight estimator) on Llama-3.1-8B-Instruct, over all sixteen
LongBench tasks at $100$ samples each.
That implementation logs, for each generation, the exact-weight
attention mass captured by its own selection, which is the same
quantity as our coverage $c$.
The audit therefore runs entirely on the method's own code,
configuration and instrumentation, and is independent of our kernels,
our scorer and our page granularity.

\begin{table}[t]
  \centering
  \small
  \setlength{\tabcolsep}{5pt}
  \caption{Realized coverage of the released Twilight implementation at its default configuration, read from its own telemetry (Llama-3.1-8B-Instruct, 16 LongBench tasks, 100 samples each). Coverage is the per-generation mean exact-weight attention mass captured by the method's own selection, and Read is the mean number of selected tokens divided by the context's word count. The configured target is $0.85$; the realized value is below it on $1600$ of $1600$ generations, in every task.}
  \label{tab:twilight-audit}
  \begin{tabular}{lcc@{\hspace{2.2em}}lcc}
    \toprule
    Task & Coverage & Read & Task & Coverage & Read \\
    \midrule
    narrativeqa    & $0.634$ & $2.1\%$ & passage\_retrieval\_en & $0.766$ & $1.2\%$ \\
    musique        & $0.719$ & $2.2\%$ & triviaqa               & $0.767$ & $1.8\%$ \\
    qmsum          & $0.726$ & $2.4\%$ & trec                   & $0.786$ & $5.4\%$ \\
    repobench-p    & $0.743$ & $5.9\%$ & multifieldqa\_en       & $0.799$ & $4.4\%$ \\
    hotpotqa       & $0.743$ & $2.5\%$ & 2wikimqa               & $0.806$ & $3.8\%$ \\
    passage\_count & $0.754$ & $1.2\%$ & lcc                    & $0.809$ & $9.3\%$ \\
    samsum         & $0.759$ & $2.9\%$ & qasper                 & $0.812$ & $4.8\%$ \\
    gov\_report    & $0.759$ & $2.8\%$ & multi\_news            & $0.815$ & $5.2\%$ \\
    \bottomrule
  \end{tabular}
\end{table}

Realized coverage falls below the configured $0.85$ target on all
$1600$ generations, with per-task means from $0.634$ to $0.815$ and a
single-generation minimum of $0.504$, so the shortfall is the normal
operating point rather than a tail event.
Output quality meanwhile stays close to the method's own dense arm
(mean $-0.35$ over the sixteen tasks under a common scorer), so nothing
in the output indicates that the target was missed.
That combination is the condition a certificate addresses.
The threshold is applied to estimated weights before the step is
executed, so the realized mass is never formed as a quantity the system
could act on, and at run time a step that met its target is
indistinguishable from one that did not.
Two scope conditions apply.
Absolute scores are not comparable across the two harnesses, so only
each system's difference from its own dense reference within its own
harness is quoted.
And the audit is at token granularity and low read ($0.9$--$4.1\%$ of context tokens); the
same pre-hoc principle at the page granularity that paged serving
stacks expose is the top-$p$ row of \cref{tab:selector-agnostic}, where
adding verification recovers $+3.67$ points.

\paragraph{Threshold-sweep telemetry.}
\cref{tab:tau-telemetry} extends the threshold sweep of
\cref{tab:tau-ablation} with the verifier's internal telemetry.
A single global $\tau$ trades traffic for coverage monotonically:
raising $\tau$ from $0.80$ to $0.97$ more than halves per-head
acceptance ($49.6\%\!\to\!19.1\%$) and grows re-scoring
($79.4\%\!\to\!98.1\%$) and effective read ($10.3\%\!\to\!13.0\%$),
while quality stays within $0.14$ points.
This is why $\tau=0.90$ is used everywhere without per-setting tuning:
the loop converts the traffic that $\tau$ controls into nearly
constant quality.

\begin{table}[t]
  \centering
  \small
\caption{Threshold sweep with verifier telemetry (three-model average; accept and re-score rates are task means). Quality moves by $0.14$ points across the full range while acceptance, re-scoring, and effective read vary substantially. $\tau=0.90$ is the deployed default at every scale.}
  \label{tab:tau-telemetry}
  \begin{tabular}{lcccc}
    \toprule
    $\tau$ & Quality (3-model) & Effective read & Per-head accept & Re-score rate \\
    \midrule
    0.80            & 40.82 & 10.3\% & 49.6\% & 79.4\% \\
    0.85            & 40.72 & 10.8\% & 44.7\% & 83.9\% \\
    0.90 (deployed) & 40.70 & 11.7\% & 37.8\% & 89.5\% \\
    0.95            & 40.84 & 12.6\% & 27.1\% & 95.4\% \\
    0.97            & 40.83 & 13.0\% & 19.1\% & 98.1\% \\
    \bottomrule
  \end{tabular}
\end{table}

For practitioners the reading is simple: $\tau$ sets the coverage
floor one insists on per head, its cost is paid in read traffic rather
than in quality, and $0.90$ is a sound default at every scale we
measured, used unchanged across three 7B architectures and at 70B.
Raising $\tau$ buys a stricter guarantee at a few points of extra
traffic; lowering it saves traffic at the same quality until the floor
becomes loose enough that recovery is no longer triggered where it is
needed.

\paragraph{Fully lightweight end-to-end.}
The selector-agnostic experiments above keep the exact scorer for the
gate's normalizer.
\cref{tab:fully-lightweight} closes that gap: a Quest min/max sketch
supplies both the page selection and the gate's all-context
normalizer, so no full-key scan runs anywhere in the pipeline,
including recovery expansion.
On Llama the fully lightweight configuration scores $40.07$ at
$12.95\%$ effective read, against $40.01$ for the same selector with an
exact normalizer (a $+0.06$ difference, paired-bootstrap 95\% CI
$[-0.73, +0.84]$): estimating the normalizer carries no discernible
quality cost, and the gated configuration retains its margin ($+2.25$ over
the ungated $37.82$, versus $+2.19$ for the exact-normalizer gate).
Exactness matters in the selector (exchanging it for the sketch costs
$1.86$ points) rather than in the certificate.
The sketch gate also fails conservatively: it accepts $24.8\%$ of head
checks against $31.1\%$ and re-scores at $99.5\%$ against $94.8\%$,
turning estimation noise into extra recovery traffic rather than silent
quality loss.
Total read traffic, including page scoring amortized over $M{=}4$, is
$\sim$0.15$\times$ dense for the fully lightweight configuration versus
$\sim$0.24$\times$ for \ssa{}-c4, since the sketch reads only per-page
min/max metadata rather than full keys.

\begin{table}[t]
  \centering
  \small
  \caption{Fully lightweight \ssa{} on Llama-3.1-8B (16 LongBench tasks, identical samples). Replacing the exact normalizer with the sketch changes quality by $+0.06$ (CI $[-0.73, +0.84]$); the remaining gap to \ssa{}-c4 is $-1.80$ (CI $[-2.82, -0.81]$) and is attributable to the selector, not the certificate. The fully lightweight path has the lowest total read traffic of any configuration measured.}
  \label{tab:fully-lightweight}
  \begin{tabular}{lccc}
    \toprule
    Configuration & Quality & Effective read & Total read incl.\ scoring \\
    \midrule
    Dense                                   & 42.57 & 100\%   & 1.00$\times$ \\
    \ssa{}-c4 (exact selector + norm.)      & 41.87 & 11.67\% & 0.24$\times$ \\
    Quest selector, exact norm.             & 40.01 & 12.36\% & 0.25$\times$ \\
    Quest selector, sketch norm.\ (lightweight) & 40.07 & 12.95\% & 0.15$\times$ \\
    \bottomrule
  \end{tabular}
\end{table}

%% file: appendix/limitations.tex
\section{Limitations and Future Directions}
\label{app:limitations}

\cyl{%
\paragraph{Scope of the median decode-step speedup.}
The median decode-step speedup that \ssa{} provides is moderate at short context
($1.76\times$ at 8K, $B{=}32$) because weight reads and non-attention
work remain a large share of the decode step at that operating point.
The advantage grows with context length and batch size and reaches
$6.21\times$ on Qwen at 128K with $B{=}8$, where KV traffic dominates
decode-step latency; Llama and Mistral reach $1.19\times$ at 128K with
$B{=}1$ (\cref{tab:longctx-throughput}).
Applications that operate primarily at short context would see limited
benefit from the current framework.

\paragraph{Theoretical coverage of reuse steps.}
The formal error guarantee (\cref{thm:error}, corollary) applies to a
single scoring step with a fixed selected set.
On reuse steps ($t \bmod M \neq 0$), the page selection and the
all-context normalizer are stale and no check runs; the check at the
next scoring step serves as an empirical safeguard rather than a formal bound.
A compositional error analysis that tracks how single-step staleness
accumulates across the reuse window is an open theoretical question.
The runtime coverage traces in \cref{fig:hybrid-coverage-trace}, and
the $32{,}768$-step trace in \cref{app:bootstrap}, indicate that selection
drift is bounded in practice, but a formal treatment would strengthen the
guarantee for deeper models where compounding has more layers to
traverse.

\paragraph{Page-level proxy fidelity.}
The deployed verifier uses a page-max proxy for coverage rather than
the exact token-level quantity.
The proxy over-accepts on all three models: in the production probe
of \cref{app:bootstrap}, $31\%$, $33\%$, and $39\%$ of proxy accepts
on Llama, Mistral, and Qwen are false accepts.
The harm concentrates on Qwen, where false accepts carry the largest
output error and cluster in late layers where the page-max saturates
under 7:1 GQA.
Token-aware or GQA-aware proxy calibration would address this drift
and is the most direct route to improving quality on architectures
with high GQA ratios.

\paragraph{Depth-dependent scheduling.}
At 70B scale the scoring interval must be reduced from $M{=}4$ to
$M{=}2$ to contain the additional per-layer drift that 80 layers
introduce.
This is currently a manual guideline based on the ablation in
\cref{tab:70b}.
An adaptive policy that sets $M$ based on observed coverage statistics
during the first few decode steps would make the framework
self-tuning, and is a natural extension of the coverage check.

\paragraph{Reasoning-length generation.}
\ssa{} targets long-context decoding, where the prompt is long but the
generated answer is short.
Long chain-of-thought reasoning, where the generated trace itself is
long, is a different regime, which we probe here to map the method's
boundary.
Transferring the deployed $M{=}4$ schedule to it unchanged is
suboptimal: on MATH-500 (R1-Distill-Llama-8B, 200 problems,
seed-paired sampling) \ssa{}-c4 reaches $64.0\%$ against dense
$80.5\%$.
The reuse window, rather than verification, accounts for the larger
part of this loss: refreshing the selection every step ($M{=}1$)
recovers two-thirds of the gap, to $75.0\%$ ($+11.0$ points over
$M{=}4$, paired-bootstrap CI $[+5.5, +16.5]$), and more than halves
both signatures of non-convergence: cap-outs fall from $31$ to $13$ of
$200$ problems and generation-length inflation from $+61\%$ to $+20\%$
above dense.
A residual $5.5$-point gap remains and the per-step schedule runs at
$0.93\times$ dense decode ($8$K, $B{=}32$), so we present
reasoning-length decoding as future work, addressed by a drift-driven
adaptive schedule that shortens the reuse window when the generation
is long (consistent with the depth-dependent guideline above), rather
than as a current capability.

\paragraph{Broader impacts.}
Reducing inference cost can improve access to long-context models.
It can also make harmful generation cheaper; deployments inherit the
underlying model's misuse, privacy, and bias risks. Attention efficiency
does not itself mitigate these risks.
}

%% file: appendix/reproducibility.tex
\section{Reproducibility}
\label{app:reproducibility}

\cyl{%
The implementation of \ssa{}, including the paged-decode pipeline,
coverage verification, hybrid scheduling, and evaluation scripts,
is prepared for public release at \url{https://github.com/DJ-Melvin/coverage-verified-sparse-attention}.
The repository contains:
(1) the Triton kernels for page scoring, sparse attention, and
coverage computation,
(2) the verify-then-correct decode loop with configurable $\tau$,
$M$, and recovery budget,
(3) evaluation harnesses for LongBench and RULER with the exact
prompt templates and sampling parameters used in this paper,
(4) launch scripts and a guide mapping every table and figure in the
main paper and appendix to its command, and
(5) pre-computed result JSON files that allow figure generation
without GPU access.
Model weights are publicly available from the respective model
providers (Meta Llama, Mistral AI, Alibaba Qwen).
LongBench and RULER use greedy decoding, which is deterministic given
the same model weights and input; the MATH-500 and stochastic-decoding
experiments use seed-paired sampling.
LongBench evaluates the first $N$ samples of each task and RULER
instances are generated from a fixed seed, so all methods see identical
inputs.
Speedup is reported as the ratio of median decode-step latencies over
the timed steps of each run; aggregate tokens/s is a different statistic
(\cref{app:longctx-table}).%
}

%% file: appendix/ruler_tables.tex
\begin{table}[t]
  \centering
  \small
  \caption{Mean RULER accuracy gap to dense (points), averaged over the six tasks and three models. The 4K--32K column covers every method on the same 72 cells and is the basis of the RULER numbers in the main text; the second column adds the 64K and 128K cells each method was run on (cell counts in parentheses).}
  \label{tab:ruler-summary}
  \begin{tabular}{@{}lcc@{}}
    \toprule
    Method & 4K--32K & All measured contexts \\
    \midrule
    \ssa{}-c4 & $-0.83$ (72) & $-0.93$ (108) \\
    \ssa{}-c8 & $-1.02$ (72) & $-1.48$ (108) \\
    SnapKV & $-16.16$ (72) & $-17.07$ (90) \\
    H2O & $-8.70$ (72) & $-9.11$ (90) \\
    AdaKV & $-9.44$ (72) & $-9.48$ (90) \\
    Fixed-10\% & $-9.35$ (72) & $-8.95$ (108) \\
    \bottomrule
  \end{tabular}
\end{table}

\begin{table}[!t]
  \centering
  \footnotesize
  \setlength{\tabcolsep}{4pt}
  \caption{RULER accuracy (\%) on Llama-3.1-8B. Through 32K each cell averages 10 trials, at 64K and 128K five; ``--'' marks a setting not run. Baselines use a $10\%$ budget.}
  \label{tab:ruler-full-llama}
  \begin{tabular}{@{}llccccccc@{}}
    \toprule
    Task & Context & Dense & \ssa{}-c4 & \ssa{}-c8 & SnapKV & H2O & AdaKV & Fixed-10\% \\
    \midrule
    S-NIAH & 4K & 100.0 & 100.0 & 100.0 & 100.0 & 100.0 & 100.0 & 100.0 \\
     & 8K & 100.0 & 100.0 & 100.0 & 100.0 & 100.0 & 100.0 & 100.0 \\
     & 16K & 100.0 & 100.0 & 100.0 & 100.0 & 100.0 & 100.0 & 100.0 \\
     & 32K & 100.0 & 100.0 & 100.0 & 100.0 & 100.0 & 100.0 & 100.0 \\
     & 64K & 100.0 & 100.0 & 100.0 & -- & -- & -- & 100.0 \\
     & 128K & 100.0 & 100.0 & 100.0 & -- & -- & -- & 100.0 \\
    \midrule
    MK-NIAH & 4K & 100.0 & 100.0 & 100.0 & 100.0 & 100.0 & 100.0 & 100.0 \\
     & 8K & 100.0 & 100.0 & 100.0 & 100.0 & 100.0 & 100.0 & 100.0 \\
     & 16K & 100.0 & 100.0 & 100.0 & 100.0 & 100.0 & 100.0 & 100.0 \\
     & 32K & 90.0 & 90.0 & 90.0 & 90.0 & 90.0 & 90.0 & 90.0 \\
     & 64K & 80.0 & 100.0 & 100.0 & -- & -- & -- & 80.0 \\
     & 128K & 100.0 & 80.0 & 80.0 & -- & -- & -- & 80.0 \\
    \midrule
    MV-NIAH & 4K & 80.0 & 70.0 & 80.0 & 70.0 & 70.0 & 80.0 & 70.0 \\
     & 8K & 60.0 & 60.0 & 60.0 & 60.0 & 60.0 & 60.0 & 60.0 \\
     & 16K & 50.0 & 50.0 & 50.0 & 50.0 & 40.0 & 60.0 & 50.0 \\
     & 32K & 60.0 & 50.0 & 50.0 & 50.0 & 50.0 & 50.0 & 50.0 \\
     & 64K & 20.0 & 20.0 & 20.0 & 20.0 & 20.0 & 20.0 & 20.0 \\
     & 128K & 40.0 & 40.0 & 40.0 & 40.0 & 40.0 & 40.0 & 40.0 \\
    \midrule
    MQ-NIAH & 4K & 100.0 & 100.0 & 96.7 & 60.0 & 90.0 & 90.0 & 93.3 \\
     & 8K & 100.0 & 100.0 & 96.7 & 66.7 & 86.7 & 90.0 & 83.3 \\
     & 16K & 100.0 & 100.0 & 100.0 & 70.0 & 90.0 & 96.7 & 93.3 \\
     & 32K & 100.0 & 96.7 & 100.0 & 60.0 & 86.7 & 80.0 & 83.3 \\
     & 64K & 93.3 & 93.3 & 100.0 & 80.0 & 86.7 & 80.0 & 86.7 \\
     & 128K & 86.7 & 86.7 & 86.7 & 80.0 & 86.7 & 86.7 & 86.7 \\
    \midrule
    VT & 4K & 70.0 & 70.0 & 80.0 & 80.0 & 80.0 & 80.0 & 80.0 \\
     & 8K & 70.0 & 70.0 & 70.0 & 70.0 & 70.0 & 70.0 & 70.0 \\
     & 16K & 80.0 & 80.0 & 80.0 & 80.0 & 80.0 & 80.0 & 80.0 \\
     & 32K & 70.0 & 70.0 & 70.0 & 70.0 & 70.0 & 70.0 & 70.0 \\
     & 64K & 40.0 & 40.0 & 40.0 & -- & -- & -- & 40.0 \\
     & 128K & 40.0 & 40.0 & 40.0 & -- & -- & -- & 40.0 \\
    \midrule
    CWE & 4K & 100.0 & 93.3 & 76.7 & 63.3 & 76.7 & 73.3 & 73.3 \\
     & 8K & 100.0 & 90.0 & 93.3 & 40.0 & 76.7 & 90.0 & 96.7 \\
     & 16K & 100.0 & 100.0 & 96.7 & 56.7 & 96.7 & 90.0 & 93.3 \\
     & 32K & 100.0 & 100.0 & 100.0 & 43.3 & 93.3 & 86.7 & 86.7 \\
     & 64K & 100.0 & 100.0 & 100.0 & 53.3 & 93.3 & 100.0 & 93.3 \\
     & 128K & 100.0 & 93.3 & 73.3 & 60.0 & 60.0 & 73.3 & 60.0 \\
    \bottomrule
  \end{tabular}
\end{table}

\begin{table}[!t]
  \centering
  \footnotesize
  \setlength{\tabcolsep}{4pt}
  \caption{RULER accuracy (\%) on Mistral-7B-v0.3. Through 32K each cell averages 10 trials, at 64K and 128K five; ``--'' marks a setting not run. Baselines use a $10\%$ budget.}
  \label{tab:ruler-full-mistral}
  \begin{tabular}{@{}llccccccc@{}}
    \toprule
    Task & Context & Dense & \ssa{}-c4 & \ssa{}-c8 & SnapKV & H2O & AdaKV & Fixed-10\% \\
    \midrule
    S-NIAH & 4K & 100.0 & 100.0 & 100.0 & 100.0 & 100.0 & 100.0 & 100.0 \\
     & 8K & 100.0 & 100.0 & 100.0 & 100.0 & 100.0 & 100.0 & 100.0 \\
     & 16K & 100.0 & 100.0 & 100.0 & 100.0 & 100.0 & 100.0 & 100.0 \\
     & 32K & 100.0 & 100.0 & 100.0 & 100.0 & 100.0 & 100.0 & 100.0 \\
     & 64K & 100.0 & 100.0 & 100.0 & -- & -- & -- & 100.0 \\
     & 128K & 100.0 & 100.0 & 100.0 & -- & -- & -- & 100.0 \\
    \midrule
    MK-NIAH & 4K & 100.0 & 100.0 & 100.0 & 90.0 & 90.0 & 90.0 & 90.0 \\
     & 8K & 100.0 & 100.0 & 100.0 & 90.0 & 90.0 & 90.0 & 90.0 \\
     & 16K & 100.0 & 100.0 & 100.0 & 90.0 & 90.0 & 90.0 & 90.0 \\
     & 32K & 100.0 & 100.0 & 100.0 & 100.0 & 100.0 & 90.0 & 90.0 \\
     & 64K & 80.0 & 60.0 & 60.0 & -- & -- & -- & 40.0 \\
     & 128K & 20.0 & 0.0 & 0.0 & -- & -- & -- & 20.0 \\
    \midrule
    MV-NIAH & 4K & 70.0 & 70.0 & 70.0 & 60.0 & 60.0 & 60.0 & 60.0 \\
     & 8K & 90.0 & 90.0 & 90.0 & 60.0 & 80.0 & 80.0 & 80.0 \\
     & 16K & 90.0 & 80.0 & 80.0 & 30.0 & 80.0 & 60.0 & 60.0 \\
     & 32K & 70.0 & 80.0 & 70.0 & 20.0 & 50.0 & 50.0 & 50.0 \\
     & 64K & 40.0 & 40.0 & 40.0 & 20.0 & 20.0 & 20.0 & 20.0 \\
     & 128K & 0.0 & 0.0 & 0.0 & 0.0 & 0.0 & 0.0 & 0.0 \\
    \midrule
    MQ-NIAH & 4K & 100.0 & 100.0 & 100.0 & 83.3 & 100.0 & 100.0 & 100.0 \\
     & 8K & 100.0 & 100.0 & 100.0 & 80.0 & 96.7 & 96.7 & 96.7 \\
     & 16K & 100.0 & 100.0 & 100.0 & 86.7 & 96.7 & 96.7 & 96.7 \\
     & 32K & 100.0 & 100.0 & 100.0 & 80.0 & 96.7 & 96.7 & 96.7 \\
     & 64K & 100.0 & 100.0 & 93.3 & 66.7 & 100.0 & 86.7 & 86.7 \\
     & 128K & 60.0 & 73.3 & 60.0 & 20.0 & 53.3 & 53.3 & 53.3 \\
    \midrule
    VT & 4K & 70.0 & 70.0 & 70.0 & 70.0 & 70.0 & 70.0 & 70.0 \\
     & 8K & 70.0 & 70.0 & 70.0 & 70.0 & 70.0 & 70.0 & 70.0 \\
     & 16K & 70.0 & 70.0 & 70.0 & 70.0 & 70.0 & 70.0 & 70.0 \\
     & 32K & 70.0 & 70.0 & 70.0 & 70.0 & 70.0 & 70.0 & 70.0 \\
     & 64K & 60.0 & 60.0 & 60.0 & -- & -- & -- & 60.0 \\
     & 128K & 20.0 & 20.0 & 20.0 & -- & -- & -- & 40.0 \\
    \midrule
    CWE & 4K & 100.0 & 96.7 & 90.0 & 100.0 & 100.0 & 100.0 & 100.0 \\
     & 8K & 100.0 & 96.7 & 96.7 & 96.7 & 100.0 & 100.0 & 100.0 \\
     & 16K & 100.0 & 96.7 & 96.7 & 86.7 & 100.0 & 100.0 & 100.0 \\
     & 32K & 100.0 & 100.0 & 96.7 & 96.7 & 100.0 & 100.0 & 100.0 \\
     & 64K & 100.0 & 100.0 & 100.0 & 100.0 & 100.0 & 100.0 & 100.0 \\
     & 128K & 93.3 & 86.7 & 86.7 & 93.3 & 86.7 & 93.3 & 93.3 \\
    \bottomrule
  \end{tabular}
\end{table}

\begin{table}[!t]
  \centering
  \footnotesize
  \setlength{\tabcolsep}{4pt}
  \caption{RULER accuracy (\%) on Qwen2.5-7B. Through 32K each cell averages 10 trials, at 64K and 128K five; ``--'' marks a setting not run. Baselines use a $10\%$ budget.}
  \label{tab:ruler-full-qwen}
  \begin{tabular}{@{}llccccccc@{}}
    \toprule
    Task & Context & Dense & \ssa{}-c4 & \ssa{}-c8 & SnapKV & H2O & AdaKV & Fixed-10\% \\
    \midrule
    S-NIAH & 4K & 100.0 & 100.0 & 100.0 & 100.0 & 100.0 & 100.0 & 100.0 \\
     & 8K & 100.0 & 100.0 & 100.0 & 100.0 & 100.0 & 100.0 & 100.0 \\
     & 16K & 100.0 & 100.0 & 100.0 & 100.0 & 100.0 & 100.0 & 100.0 \\
     & 32K & 100.0 & 100.0 & 100.0 & 100.0 & 100.0 & 100.0 & 100.0 \\
     & 64K & 100.0 & 100.0 & 100.0 & -- & -- & -- & 100.0 \\
     & 128K & 20.0 & 20.0 & 20.0 & -- & -- & -- & 20.0 \\
    \midrule
    MK-NIAH & 4K & 100.0 & 100.0 & 100.0 & 70.0 & 70.0 & 70.0 & 70.0 \\
     & 8K & 100.0 & 100.0 & 100.0 & 70.0 & 70.0 & 70.0 & 70.0 \\
     & 16K & 90.0 & 100.0 & 100.0 & 70.0 & 70.0 & 70.0 & 70.0 \\
     & 32K & 100.0 & 100.0 & 100.0 & 70.0 & 70.0 & 70.0 & 70.0 \\
     & 64K & 60.0 & 60.0 & 60.0 & -- & -- & -- & 40.0 \\
     & 128K & 20.0 & 20.0 & 20.0 & -- & -- & -- & 0.0 \\
    \midrule
    MV-NIAH & 4K & 100.0 & 100.0 & 100.0 & 30.0 & 60.0 & 40.0 & 60.0 \\
     & 8K & 90.0 & 90.0 & 90.0 & 20.0 & 70.0 & 60.0 & 60.0 \\
     & 16K & 100.0 & 90.0 & 100.0 & 30.0 & 80.0 & 70.0 & 70.0 \\
     & 32K & 80.0 & 80.0 & 80.0 & 40.0 & 60.0 & 60.0 & 50.0 \\
     & 64K & 80.0 & 80.0 & 80.0 & 20.0 & 60.0 & 80.0 & 60.0 \\
     & 128K & 60.0 & 60.0 & 60.0 & 60.0 & 60.0 & 60.0 & 60.0 \\
    \midrule
    MQ-NIAH & 4K & 100.0 & 100.0 & 100.0 & 56.7 & 63.3 & 53.3 & 50.0 \\
     & 8K & 100.0 & 100.0 & 100.0 & 46.7 & 50.0 & 43.3 & 50.0 \\
     & 16K & 100.0 & 100.0 & 100.0 & 60.0 & 40.0 & 43.3 & 40.0 \\
     & 32K & 100.0 & 96.7 & 96.7 & 60.0 & 53.3 & 53.3 & 63.3 \\
     & 64K & 100.0 & 93.3 & 93.3 & 33.3 & 53.3 & 53.3 & 46.7 \\
     & 128K & 66.7 & 60.0 & 53.3 & 40.0 & 40.0 & 46.7 & 33.3 \\
    \midrule
    VT & 4K & 70.0 & 70.0 & 70.0 & 70.0 & 70.0 & 70.0 & 70.0 \\
     & 8K & 70.0 & 70.0 & 70.0 & 70.0 & 70.0 & 70.0 & 70.0 \\
     & 16K & 70.0 & 70.0 & 70.0 & 70.0 & 70.0 & 70.0 & 70.0 \\
     & 32K & 70.0 & 70.0 & 70.0 & 70.0 & 70.0 & 70.0 & 70.0 \\
     & 64K & 40.0 & 40.0 & 40.0 & -- & -- & -- & 40.0 \\
     & 128K & 20.0 & 20.0 & 20.0 & -- & -- & -- & 20.0 \\
    \midrule
    CWE & 4K & 100.0 & 93.3 & 90.0 & 96.7 & 93.3 & 96.7 & 96.7 \\
     & 8K & 100.0 & 100.0 & 100.0 & 100.0 & 100.0 & 96.7 & 96.7 \\
     & 16K & 100.0 & 100.0 & 100.0 & 96.7 & 96.7 & 96.7 & 96.7 \\
     & 32K & 100.0 & 100.0 & 100.0 & 100.0 & 100.0 & 100.0 & 100.0 \\
     & 64K & 100.0 & 100.0 & 100.0 & 93.3 & 93.3 & 93.3 & 93.3 \\
     & 128K & 86.7 & 100.0 & 93.3 & 73.3 & 80.0 & 66.7 & 80.0 \\
    \bottomrule
  \end{tabular}
\end{table}